%% file: main.tex
\documentclass[letterpaper]{article} 
\usepackage{aaai2026}  
\usepackage{amssymb} 
\usepackage{amsmath}
\usepackage{comment}
\usepackage{booktabs}
\usepackage{multirow}
\usepackage[percent]{overpic}

\usepackage{subcaption}
\usepackage[above,below]{placeins}

\usepackage{xcolor}    
\usepackage{soul}      

\usepackage{times}  
\usepackage{helvet}  
\usepackage{courier}  
\usepackage[hyphens]{url}  
\usepackage{graphicx} 
\usepackage{natbib}  
\usepackage{caption} 
\usepackage{algorithm}
\usepackage{algorithmic}

\usepackage{tikz}
\usetikzlibrary{shapes.geometric, arrows.meta, positioning, fit}

\usepackage{newfloat}
\usepackage{listings}

\newcommand{\bX}{\ensuremath{\mathbf{X}}}
\newcommand{\E}{\ensuremath{\mathbb{E}}}

\newtheorem{proposition}{Proposition}

\DeclareCaptionStyle{ruled}{labelfont=normalfont,labelsep=colon,strut=off} 
\floatstyle{ruled}
\newfloat{listing}{tb}{lst}{}
\floatname{listing}{Listing}
\title{Debiasing Machine Learning Predictions for Causal Inference Without Additional Ground Truth Data: ``One Map, Many Trials'' in Satellite-Driven Poverty Analysis
}

\author{
    Markus B. Pettersson\textsuperscript{\rm 1},
    Connor Thomas Jerzak\textsuperscript{\rm 2},
    Adel Daoud\textsuperscript{\rm 1,3}
}
\affiliations{
    \textsuperscript{\rm 1}Data Science \& AI Division, Department of Computer Science, Chalmers University of Technology, Gothenburg, Sweden\\
    \textsuperscript{\rm 2}Department of Government, The University of Texas at Austin, Austin, TX, USA\\
    \textsuperscript{\rm 3}Institute for Analytical Sociology, Linköping University, Norrköping, Sweden\\
    {markus.pettersson@chalmers.se}, {connor.jerzak@austin.utexas.edu}, {adel.daoud@liu.se}
}

\usepackage{bm}      

\usetikzlibrary{
    arrows.meta,                
    positioning,                
    calc,                       
    fit,                        
    shapes.geometric            
}

\begin{document}

\maketitle

\begin{abstract}
Machine learning models trained on Earth observation data, such as satellite imagery, have demonstrated significant promise in predicting household-level wealth indices, enabling the creation of high-resolution wealth maps that can be leveraged across multiple causal trials while addressing chronic data scarcity in global development research. However, because standard training objectives prioritize overall predictive accuracy, these predictions often suffer from shrinkage toward the mean, leading to attenuated estimates of causal treatment effects and limiting their utility in policy evaluations. Existing debiasing methods, such as Prediction-Powered Inference (PPI), can handle this attenuation bias but require additional fresh ground-truth data at the downstream stage of causal inference, which restricts their applicability in data-scarce environments. 
We introduce and evaluate two post-hoc correction methods—Linear Calibration Correction (LCC) and a Tweedie's correction approach—that substantially reduce shrinkage-induced prediction bias without relying on newly collected labeled data. LCC applies a simple linear transformation estimated on a held-out calibration split; Tweedie's method locally de-shrink predictions using density score estimates and a noise scale learned upstream. We provide practical diagnostics for \emph{when} a correction is warranted and discuss practical limitations. 
Across analytical results, simulations, and experiments with Demographic and Health Surveys (DHS) data, both approaches reduce attenuation; Tweedie's correction yields nearly unbiased treatment-effect estimates, enabling a ``one map, many trials'' paradigm. Although we demonstrate on EO-ML wealth mapping, the methods are not geospatial-specific: they apply to any setting where imputed outcomes are reused downstream (e.g., pollution indices, population density, or LLM-derived indicators).
\end{abstract}

\section*{Introduction}

Machine learning (ML) models trained on Earth observation (EO) data have recently demonstrated impressive performance in predicting household-level wealth indices, such as the International Wealth Index (IWI), with reported $R^2$ values approaching 0.80 \citep{pettersson2023time,jean2016combining,daoud2023using,chi2022microestimates}. These EO-ML models offer a much-needed solution to longstanding data scarcity issues in global development, enabling researchers to generate high-resolution, retrospective ``wealth maps'' in data-poor regions. Indeed, two recurring downstream uses are: (i) evaluating aid or infrastructure programs (\emph{e.g.}, differences in post-intervention wealth across treated vs. control regions), and (ii) tracking regional poverty trends over time. 

\begin{figure}[htb]
    \centering
    \includegraphics[width=1\linewidth]{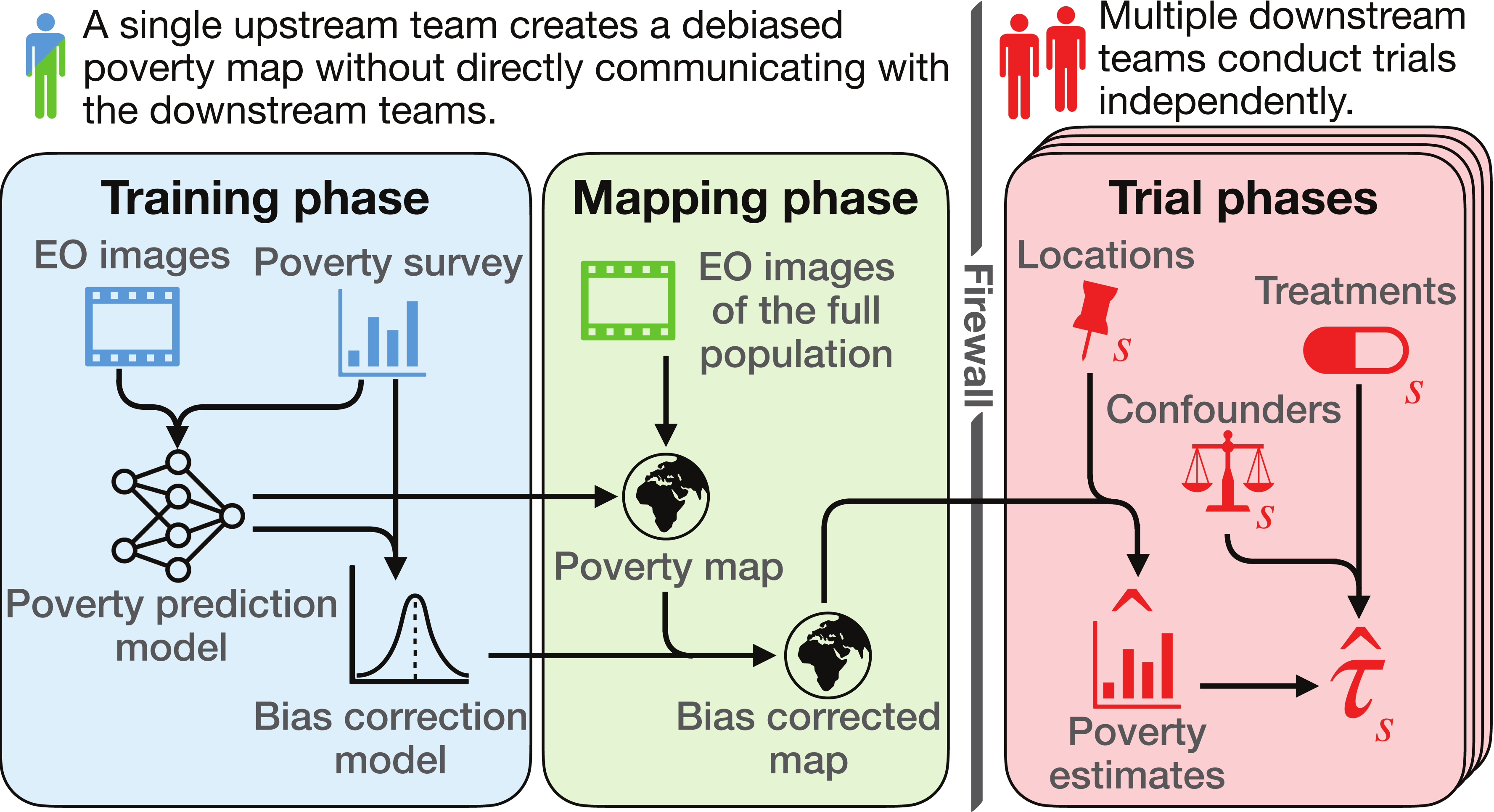}
    \caption{
    Pipeline for causal inference using ML predictions without fresh ground-truth data. Upstream teams train and calibrate once; downstream teams reuse the same map across multiple trials. 
    }
    \label{fig:flowchart}
\end{figure}

But while these predictions are promising in the EO-ML upstream task, a key question remains \citep{jerzak2023integrating,jerzak2023image}: the value of EO-ML-derived data hinges on whether domain researchers, in the downstream phase, can use these data to estimate causal effects for their specific application areas, from evaluating applied environmental to social science questions \citep{daoudStatisticalModelingThree2023}. 
However, even under ideal conditions, the predictions of EO-ML models tend to have lower variance than the target values, leading to attenuated treatment effects and, thus, potentially misleading conclusions \citep{berglund2012regression}. When upstream predictions shrink toward the global mean, estimated effects are biased toward zero, increasing Type-II error: a true 5\% effect can appear as only 2–3\%, inviting ``no impact'' conclusions despite meaningful gains. Figure~\ref{fig:flowchart} provides a pipeline overview linking the upstream map to multiple downstream trials without requiring new labels.

Recent research has developed methods that can debias upstream machine-learning model predictions, but they require either fresh data in the downstream stage \citep{olofsson2013making,dsl_original,lu2024quantifying} or changing the way the model is trained \citep{ratledge_using_2022}. For example, relying on such fresh data, \citet{angelopoulos2023prediction} propose Prediction-Powered Inference (PPI), a method that corrects any ML model for downstream inference. Similarly, \citet{dsl_original} achieve the same goal but with better sample efficiency \citep{depieuchon2025benchmarkingdebiasingmethodsllmbased}, but it also requires knowing the probability of observing a sample point. While these methods offer a path to integrate upstream and downstream analyses, their success depends on the availability of newly collected treatment-predictor-outcome tuples in the downstream causal step or knowledge of the data labeling process \citep{dsl_original}. Often, such data are prohibitively costly or unethical to collect.
 
\citet{ratledge_using_2022} address the challenge of attenuation bias and the unavailability of data in the downstream by adjusting the model's loss function instead. During training, they penalize quintile-level discrepancies between observed and predicted means, encouraging predictions to match the true outcome distribution more closely across its support. Although their solutions do not need additional downstream data, they require modifying the training pipeline, which is computationally expensive and can degrade predictive performance.

Thus, to enable broader applicability of ML models, the upstream team needs to create a data product that is agnostic to downstream use, requires no downstream-fresh data, produces unbiased causal estimates (due to attenuation), and is computationally efficient \citep{daoudStatisticalModelingThree2023}.

We address this gap by proposing a linear calibration method and a family of Tweedie corrections, a novel approach that reduces prediction bias without requiring fresh labeled data. We evaluate these two approaches against three others: Prediction-Powered Inference (PPI), the loss-adjustment approach of \citet{ratledge_using_2022}, and naive (uncorrected) predictions. The methods introduced here pave the way for the same wealth map to be reused across multiple trials without needing to refit a model for each trial. This also removes the requirement for interaction between the upstream machine-learning team creating the data product and any other domain-knowledge team in the downstream phase. Hence the concept: \textit{one map, many trials}. We derive analytical proofs and demonstrate the empirical performance of all methods in both simulations and real-world applications, using data from the African Demographic and Health Survey (DHS). 

We make these corrections accessible via an open-source software package called \texttt{unshrink} (see Appendix F); although developed for EO-ML wealth maps, these same attenuation and remedies arise broadly whenever surrogate outcomes replace direct measurements, including in environmental (air pollution, urban heat), demographic (population density), and text/LLM-annotated constructs.

\section{Problem Setup \& Related Work}

Statistical predictions can systematically shrink estimates toward the central modes of the training distribution \citep{ting2024machine}. This phenomenon is rooted in optimal error-minimization estimation (cf. Stein’s paradox), where a model ``hedges'' predictions in regions of greater uncertainty. When facing noise or limited training data, the optimal strategy, in terms of minimizing residuals, is to pull predictions toward denser regions of the training data. For instance, consider an image that appears with an IWI of 80, but the model has observed this outcome only once, compared to thousands of observations with an IWI of around 70. A perfect prediction would be 80 for this data point, but the model will pull this data point's prediction towards 70 instead to hedge against the risk of overcommitting to a rare value. 

\begin{figure}[ht]
    \centering
        \includegraphics[width=1\linewidth]{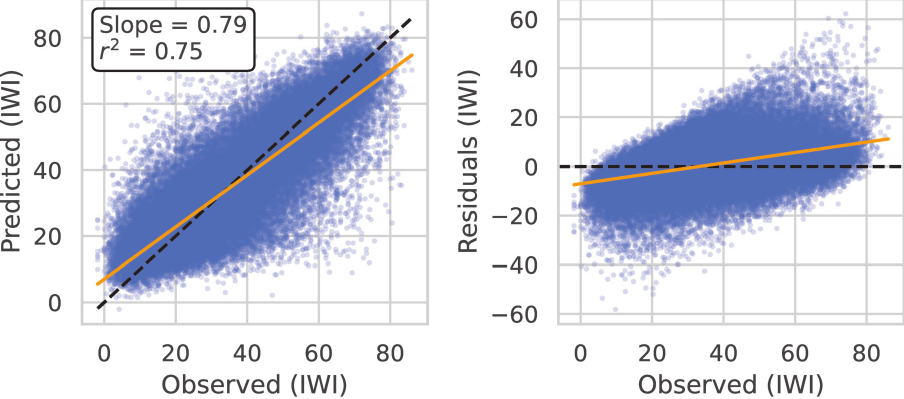}
    \caption{
    The left is a calibration plot of predicted versus true IWI values at held-out DHS clusters. The orange line shows that the predictions exhibit attenuation bias, resulting in overestimation of wealth for poorer locations and underestimation for wealthier ones. The dashed lines represent the ideal relationship. The right plot shows the residuals.}
    \label{fig:cluster_level_preds}
\end{figure}

When such biased predictions replace true outcomes in downstream causal analyses, the resulting regression-to-the-mean induces attenuation bias, reducing observed treatment group differences even under randomization \citep{shu2019causal}. The difference in mean predicted outcomes between treated and control units shrinks both group means toward the global average, thereby attenuating the estimated effect. In our held-out DHS clusters, the calibration curve exhibits this shrinkage pattern—overpredicting poor locations and underpredicting wealthy ones (Figure~\ref{fig:cluster_level_preds}).

Regression-to-the-mean bias is particularly acute in upstream tasks with scarce ground truth (due to finite samples) or in high-dimensional input data (due to the curse of dimensionality), forcing greater reliance on error-prone predictions---something especially pertinent in development economics, political science, and global health contexts---where household data is often expensive or unavailable—to increasingly substitute upstream predictions \citep{kino2021scoping,kakooei2024mapping,daoudImpactAusterityChildren2024}. Moreover, high out-of-sample $R^2$ does not eliminate this bias: models can achieve strong accuracy yet yield attenuated causal estimates \citep{ting2024machinelearningmodelssystematically}.

Existing remedies are PPI, which subtracts a rectifier estimated from fresh ground-truth pairs at the causal-inference stage \citep{angelopoulos2023prediction}, or loss-function adjustments that penalize quintile-level errors \citep{ratledge_using_2022}. However, these approaches require that the upstream team adapts the data for downstream analysis, slowing or limiting altogether ``one map, many trials'' efforts. 

To address these limitations, we propose a Tweedie-inspired approach. This Tweedie approach debiases the shrinkage effect in ML-predicted outcomes, enabling downstream researchers to accurately recover causal effects from the upstream data product. Our Tweedie framework treats the ML model as a black box: we assume access only to its out‑of‑sample predictions and do not rely on retraining or modifying the model. Under this setting, we demonstrate that shrinkage can be reversed, either by learning a global calibration correction from held-out labeled data or by estimating a local correction using ideas from Tweedie's formula, initially described by Robbins and Tweedie in 1956 in the context of empirical Bayes estimation of means under Gaussian observation noise. Both strategies aim to recover unbiased treatment effect estimates from biased predictions, without requiring fresh ground-truth labels at the time of intervention. We benchmark our Tweedie adaptation to existing methods, which are compared in Table \ref{tab:debiasing-methods} based on their associated assumptions and data/training requirements.\footnote{
As previously mentioned, the only method we are unable to deploy in our setting is the DSL method, as the probability of DHS samples depends on unknown country-level DHS sampling strategies (probabilities not publicly provided at the unit level).
} In short, Tweedie’s formula is canonical; we adapt it to post-hoc, local de-shrinkage of surrogate outcomes used for causal estimation without new labels,  estimating the density score and noise scale from upstream data and applying correction downstream at analysis time.

\noindent \textbf{Setup.} Our pipeline assumes access to two datasets. First, from upstream, a labeled dataset with training points (with indices in $\mathcal{I}_{\textrm{Train}}$): $D_{\text{Train}} = \{(\bX_i, Y_i): i\in\mathcal{I}_{\textrm{Train}}\},$
where $\bX_i \in \mathcal{X}$ denotes covariates (e.g., satellite imagery) and $Y_i$ is  ground-truth outcome (e.g.,  poverty level from the DHS).

Next, in the downstream phase, we assume that different teams have assembled collections of trial datasets indexed by \(s\), corresponding to different randomized controlled trials, quasi-experiments, or geographic partitions, with a specific treatment (or \underline{a}ction) of interest indexed by treatment setting:
\(
D_{\text{Trial $s$}} = \{(\bX_i, \hat{Y}_i,  A_{is}): i\in \mathcal{I}_{\textrm{Trial $s$}}, s \in \mathcal{S} \}.
\)
The data product associated with the outcome is population-wide (e.g., all-Africa), but investigators may want to use that data in multiple causal settings, \(s\), each with a different treatment,  \(A_{is} \in \{0,1\}\). The outcome \(Y_i\) is not directly observed for all $i$. Instead, inference will need to take place through the noisy variable, $\hat{Y}_i$. We henceforth remove dependence on $s$ for simplicity.

A key idea here is that the upstream team---a machine-learning team---creates an \textit{agnostic} dataset \citep{grimmer2022text}, meaning that this team's data product should produce unbiased causal estimates for any number of applications (trials) for a variety of (social science) teams downstream. This setup implies that the upstream team versus the downstream teams should be able to execute on their respective research goals without communication---hence the firewall between the two phases.  

\section{Debiasing Methods}

In this section, we present a suite of methods designed to debias treatment effect estimates that rely on ML-predicted outcomes. Our overarching goal is to mitigate regression-to-the-mean dynamics in the downstream analysis, improving estimates of causal effects or population-level statistics. We first more formally describe two existing solutions, \textit{PPI} and the method of \citet{ratledge_using_2022}, before introducing two no-new-data debiasing approaches---(1) Linear Calibration Correction (LCC) and (2) Tweedie’s Corrections---that aim to correct shrinkage bias without requiring fresh labels in the causal step.

\subsection*{Existing Debiasing Methods (PPI, Ratledge, DSL)}

PPI proposes a framework for valid statistical inference when a large set of unlabeled observations is paired with a predictive model’s outputs \citep{angelopoulos2023prediction}. Concretely, letting $\theta^*$ be the target parameter (e.g., a population mean) and $f$ denote the model that maps covariates $\bX_i$ to predicted outcomes $f(\bX_i)$, PPI forms an estimator by adding a \emph{rectifier} term---an adjustment for the average discrepancy between $Y_i$ and $f(\bX_i)$ observed in a smaller, labeled dataset. An estimate for $\theta^*$ is computed using an ``imputed'' version based on unlabeled data ($\{\bX_i\}_{i={n+1}}^N$), then corrected by subtracting the bias estimated on the labeled subset. For mean estimates, this adjustment is:  
\[
\hat{\theta}^{\textrm{PPI}} = \frac{1}{N-n} \sum_{i={n+1}}^N f(\bX_i) - \frac{1}{n} \sum_{i=1}^n \left( f(\bX_i) - Y_i \right)
\]
Because the rectifier is estimated on ground-truth pairs $\{\bX_i, Y_i\}_{i=1}^n$, the procedure maintains valid coverage for $\theta^*$ in large samples, provided that the predictive model $f$ is fixed (i.e., not retrained on the labeled data). Unlike our approach, however, PPI crucially relies on gathering labeled observations in the downstream phase. Such labels come from high-quality surveys coordinated by USAID across the African continent and cost millions of US dollars to execute. Given the 2025 funding cuts to USAID, it is unlikely that new and fresh DHS data points will be collected. Hence, there is an acute need for unbiased data creation, which should be fully handled in the upstream phase. We will nonetheless include PPI correction in our analysis with the goal of showing {\it what would have happened to our estimates had fresh data actually been available downstream.} 

Another approach to remedy attenuation bias is to adjust the loss function: \citet{ratledge_using_2022} propose a specialized upstream training procedure that directly penalizes quintile-level discrepancies between observed and predicted means, thereby encouraging \(\hat{Y}_i\) to match the true distribution of \(Y_i\) more closely across the support of the outcome distribution. Specifically, the training objective augments the mean squared error loss with an $\ell_2$ penalty and a quintile‑bias term:
\[
\mathcal{L}(\theta) \;=\; \mathrm{MSE}(\theta) \;+\; \lambda_r \,\|\theta\|_2^2 \;+\; \lambda_b \,\hat{E}_b(\theta),
\]
where MSE($\theta$) is the mean squared error given parameters,  $\theta$,
$\lambda_r \,\|\theta\|_2^2$ is a regularization term, and
\(\lambda_b \hat{E}_b(\theta)\) is the quintile-specific bias loss, with $\hat{E}_b(\theta)=\max_{j} \{\hat{B}_j(\hat f_\theta)^2\}$ and $\hat{B}_j(\hat f_\theta)=\mathbb{E}[\hat f_\theta(\bX_i)-Y_i\mid Y_i\in Q_j]$ penalizing quintile‑level discrepancies (where quantiles are indexed by $j$). Here, $\lambda_r$ and $\lambda_b$ denote the weighting terms for the regularization and quantile-specific loss, respectively.\footnote{
In our experiments, we use AdamW with weight decay for $\ell_2$ regularization; to avoid double‑counting, we set $\lambda_r=0$ whenever weight decay is non‑zero.
}


Through the quintile-specific bias loss, this approach reduces the likelihood that group-level estimates of causal effects are attenuated when \(\hat{Y}_i\) is used in downstream analyses by directly operating on model parameters, $\theta$. In contrast, the two new methods introduced next operate as post-hoc corrections on out-of-sample predictions from a black-box model, requiring no specialized retraining or customized loss function that might reduce predictive performance.

\subsection{Linear Calibration Correction (LCC)}\label{s:LinearCalibration}
For LCC, we approximate the relationship between EO-ML predictions and ground-truth wealth on a held-out calibration set by a linear transformation:
\[
\mathbb{E}[\hat Y_i \mid Y_i, \, A_i] = \mathbb{E}[\hat Y_i \mid Y_i] = k\,Y_i + m,
k>0,\ m\in\mathbb{R}. (A1)
\]
When $0 < k \leq 1$, shrinkage toward the mean is indicated; $m \in \mathbb{R}$ is the intercept, representing the mean fixed bias. The first equality essentially indicates that there is now conditional independence of $\hat Y_i$ from group/treatment variable $A_i\in\{0,1\}$ given $Y_i$; the second equality indicates a global affine (linear) calibration of predictions to truth, i.e., \(\mathbb{E}[\hat Y_i\mid Y_i]=kY_i+m\) with constant slope and intercept. This linearity might not always be empirically satisfied, but appears to occur approximately in our data (see, e.g., Figure \ref{fig:survey_estimates}). When these predictions are used in place of true outcomes to estimate causal effects, this shrinkage attenuates estimated treatment effects.

Let $\tau = \mathbb{E}[Y_i \mid A_{i}=1] - \mathbb{E}[Y_i \mid A_{i}=0]$ denote the true Average Treatment Effect (ATE). Using $\hat{Y}_i$ instead of $Y_i$ for ATE estimation yields the bias shown in Proposition \ref{prop:attenuation}.
\begin{proposition}[Attenuation bias under linear shrinkage]\label{prop:attenuation}
Let $Y_i$ be the true outcome and $\hat Y_i$ an ML prediction obeying the linear calibration model (A1) with $k$ and $m$ constants. Let $A_i\in\{0,1\}$ denote a binary treatment that is randomly assigned ($A_i \perp Y_i(a)$ for $a\in\mathcal{A}$). Define the true average treatment effect (ATE)  
$\tau := \mathbb{E}[Y_i(1)]-\mathbb{E}[Y_i(0)]$  and the estimate, $\hat{\tau}$, from the difference in predicted expected outcomes:
\[
\hat{\tau} := \mathbb{E}[\hat{Y}_i\mid A_i=1]-\mathbb{E}[\hat{Y}_i\mid A_i=0].
\]
Then \(\mathbb{E}[\hat\tau] \;=\; k\,\tau.\)
That is, the ATE estimated on the predictions is multiplied by the shrinkage factor~$k$ and is therefore attenuated toward~$0$ whenever $k<1$.
\end{proposition}

\noindent \textit{Proof.}
Take expectations and use linearity:
\begin{align*}
\hat{\tau} &= \mathbb{E}[\hat{Y}_i \mid A_i=1] - \mathbb{E}[\hat{Y}_i \mid A_i=0]
\\ &= \mathbb{E}[ \mathbb{E}[\hat{Y}_i \mid Y_i, A_i=1] \mid A_i=1 ] 
\\ &\qquad- \mathbb{E}[ \mathbb{E}[\hat{Y}_i \mid Y_i, A_i=0] \mid A_i=0 ]
\\ &= \mathbb{E}[ k Y_i + m \mid A_i=1 ] - \mathbb{E}[ k Y_i + m \mid A_i=0 ]
\\ &= k\{\mathbb{E}[Y_i \mid A_i=1] - \mathbb{E}[Y_i \mid A_i=0]\} = k \tau.
\end{align*}
Replacing observed with potential outcomes follows from randomization; ATE estimates with $\hat{Y}_i$ are biased downward by a factor of $k$. 

Now, to adjust for this attenuation bias, we will estimate \(k\) and \(m\) on a held-out calibration set from the training data by regressing \(\hat{Y}_i\) on \(Y_i\) (i.e., fitting the linear model with functional form, \(\hat{Y}_i = \hat{k} Y_i + \hat{m} + e_i\), where \(e_i\) are mean 0 residuals). We then invert this fitted relationship (solving for $Y_i$, which gives \(Y_i = \frac{(\hat{Y}_i - \hat{m} - e_i)}{\hat{k}}\)) to obtain a calibrated estimate, given by \(\hat{Y}^L_i = \frac{\hat{Y}_i - \hat{m}}{\hat{k}}\), assuming conditional mean zero residuals \(e_i\). Here, $\hat{Y}^L_i$ reflects the calibrated estimate of the true outcome, \(Y_i\). This calibrated estimator restores the scale of the outcome variable, yielding Proposition \ref{prop:lcc}:

\begin{proposition}[Consistency of LCC correction]\label{prop:lcc}
Under the conditions of Proposition~\ref{prop:attenuation}, estimate $(k,m)$ on an independent calibration sample and set\footnote{To be clear, $(\hat k,\hat m)$ are estimated on a random validation-style calibration split carved from the upstream labeled data.}: 
\[
\hat{Y}^L_i :=(\hat{Y}_i - \hat{m})/\hat k;\;
\hat{\tau}^L := {\mathbb{E}}[\hat{Y}^L_i|A_i=1]-{\mathbb{E}}[\hat{Y}^L_i | A_i=0].
\]
If $(\hat k,\hat m)\xrightarrow{p}(k,m)$, and $k>0$, then $\hat{\tau}^L\xrightarrow{p}\tau$; i.e., the linear calibration correction removes the attenuation bias.
\end{proposition}
\noindent \textit{Proof sketch.}
By Slutsky's theorem, since $(\hat k, \hat m) \xrightarrow{p} (k, m)$, it follows that: 
\begin{align*} 
\hat{Y}_i^L &= [ [k Y_i + m + e_i] - \widehat{m}] / \hat{k}
\\ &= \underbrace{(k/\hat{k}) Y_i}_{\color{gray}\to Y_i} + \underbrace{(m - \hat{m})/\hat{k}}_{\color{gray} \to 0} + \underbrace{e_i / \hat{k}}_{\color{gray} e_i / k} \to Y_i + e_i / k.
\end{align*} 
Since $\mathbb{E}[e_i|A_i] = 0$ under the conditional mean 0 assumption outlined earlier,
\begin{align*} 
\hat{\tau}^L &\to \mathbb{E}[ 
Y_i + e_i / k \mid A_i = 1
] - \mathbb{E}[ 
Y_i + e_i / k \mid A_i = 0
] 
\\&= \mathbb{E}[ 
Y_i \mid A_i = 1
] - \mathbb{E}[ 
Y_i \mid A_i = 0] = \tau. 
\end{align*} 
A drawback is that we assume a linear model for the shrinkage mechanism, which enters through an assumed functional form for the expected value of $\hat{Y}_i$ given $Y_i$ (cf. A1); estimating $\hat{m}$ and $\hat{k}$ introduces another potential source of error.

\input{./tweedie_main.tex}

\section{Experiments with Simulated Data}

\begin{figure}[htb]
    \centering
    \includegraphics[width=1\linewidth]{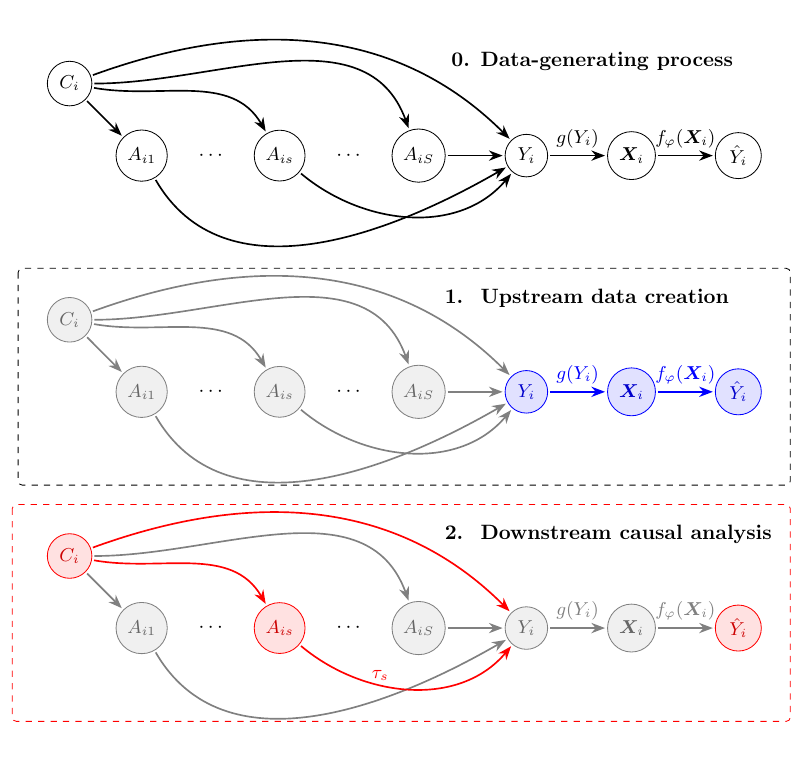}
\caption{
A panel illustrating the problem setup.\\
      \textbf{0.} Reality (data generation) shows the complete causal DAG.\\
      \textbf{1.} \textcolor{blue}{Upstream} phase: $Y_i$, $\mathbf{X}_i$, and $\hat Y_i$ are observed (\textcolor{blue}{blue}).\\
      \textbf{2.} \textcolor{red}{Downstream} phase: $A_{is}$, $C_{i}$, and $\hat Y_{i}$ are available (\textcolor{red}{red}.)
      \label{fig:DAG}
}
\end{figure}

\paragraph{Simulation Design.} Our simulation study benchmarks all debiasing methods and monitors the upstream and downstream analyses while maintaining the firewall between them. The simulation uses a canonical directed acyclic graph (DAG), shown in Figure \ref{fig:DAG}. It is \textit{canonical} in the sense that the EO-ML literature tends to assume this structure \citep{sakamoto2024scoping}, while  illustrating the various research phases and corresponding data needs. 

In phase 0, the data-generation process (approximating \textit{reality}) unfolds in the direction of the arrows. In the upstream phase, only the blue-nodes data are available, with finite samples of $Y_i$, to produce $\hat{Y_i}$ for the entire population $N$. In the downstream, only red nodes are available: $\hat{Y_i}$ is now available for $N$, jointly with $C_i$ and $A_{is}$. They enable inference on the estimand $\tau_s$. (We suppress the index $s$ henceforth.) Appendix D delineates our simulation further. 

\paragraph{Simulation Results.} We focus on two key objectives in the downstream task: minimizing error and minimizing causal bias. To assess error, we measure how closely the predicted values match the true values of $\tau$, using Mean Absolute Error (MAE). A lower MAE indicates a more accurate recovery of the true treatment effects. To assess bias, we examine the calibration slope from regressing the predicted effects $\hat{\tau}$ on the true $\tau$ values and analyzing the calibration slope. If the estimates are unbiased, the slope should be close to one. Intuitively, a slope of one means predicted and true scales align (no global shrinkage) so downstream treatment effects are not attenuated purely by scaling. Deviations from one indicate systematic over-/underestimation across the range of $\tau$. We report both the slope and its standard error, checking whether the 95\% confidence interval includes one. Figure \ref{fig:simulated_debiased_ate_estimates} visualizes these slope results. 

Results (Table \ref{tab:SimPerformance}) show that Tweedie's correction achieves the lowest MAE and yields slope estimates whose confidence interval contains one, indicating no detectable attenuation bias. In contrast, both the Ratledge and naive approaches yield calibration slopes significantly below one, indicating systematic underestimation of effects.

\begin{table}[htb]
        \centering\small
        \begin{tabular}{lccc}
          \hline \footnotesize
          \textbf{Method} & \textbf{MAE} & \textbf{Slope $\pm$ SE} & \textbf{1 $\in$ 95\% CI?} \\ \hline
          Tweedie's  & \textbf{0.04} & \textbf{0.995 $\pm$ 0.006} & $\checkmark$ \\
          LCC        & 0.05          & 1.008 $\pm$ 0.007          & $\checkmark$         \\
          PPI (10\%) & 0.19          & 0.985 $\pm$ 0.028          & $\checkmark$         \\
          Ratledge   & 0.37          & 0.641 $\pm$ 0.024          & $\times$        \\
          Naive      & 0.48          & 0.535 $\pm$ 0.004          & $\times$       \\ \hline
        \end{tabular}
        \caption{Performance on simulated treatment-effect estimation. Tweedie's correction yields the lowest error and an unbiased slope; Ratledge and the naive approach show substantial attenuation bias.}\label{tab:SimPerformance}
\end{table}

\section{Experiments with Real Data}

In this section, we conduct two types of downstream-real experiments. First, we will evaluate how well our debiasing methods succeed in descriptive tasks, especially those involving group aggregation. Second, we will apply our procedure to causal tasks. These two tasks cover the main use cases of our EO-ML-driven poverty data (see Figure~\ref{fig:combined_experiment}).

The full dataset consists of \( n \approx\) 69,000 geo-located survey clusters from the Demographic and Health Surveys (DHS) across 30 African countries (2009–2020), where each cluster comprises 20–30 households sampled via stratified two-stage design (primary sampling units selected with probability proportional to population size, followed by random household selection). We partitioned the data into five random folds for cross-validation (CV). For each run, four folds were used as upstream data (three for training, one for calibration): \( D_{\text{Train}} = \{( \mathbf{X}_i, Y_i ): i \in \mathcal{I}_{\text{Train}} \} \). Downstream trial datasets \( D_{\text{Trial } s} \) were drawn from the remaining, fifth fold and geo-referenced aid interventions, $s$, ensuring no overlap with training data. 

Here, \( Y_i \) denotes the International Wealth Index (IWI), a continuous asset-based wealth score (0–100) aggregated at the cluster level to mitigate noise \citep{burkeUsingSatelliteImagery2021,pmlr-v275-zhu25a}. The covariate array \( \mathbf{X}_i \) consists of Landsat satellite imagery (30 m resolution, six multispectral bands) centered on each DHS cluster (6.72 km × 6.72 km tiles), processed by taking the three-year median over cloud-free pixels leading up to the survey date.

Two models were fitted for each fold, one using MSE and one using the loss introduced by \citet{ratledge_using_2022}. The models shared the same ResNet-50 architecture and were initialized with weights pretrained on the SSL4EO-L dataset \citep{stewart2023ssl4eoldatasetsfoundationmodels}. For more details about model training, see Appendix B. The resulting MSE models all have an $R^2$ score between 0.74 and 0.76 on their hold-out fold, and an average slope of 0.79 . Their combined predictions can be seen in Figure \ref{fig:cluster_level_preds}. The Ratledge models have a slope of 0.99, but at the cost of reduced $R^2$ scores between 0.67 and 0.70.

\paragraph{Estimating Region Means with Bias Adjustment.} In our descriptive task, we estimate conditional means, averaging wealth by local second-level administrative region (ADM2), roughly corresponding to municipalities or counties. We group the clusters in the held-out downstream fold by ADM2 region and compare the mean observed IWI with the predicted IWI with and without the different methods of bias correction. These quantities are subject to the same shrinkage bias as causal estimates, and our framework applies equally well in this setting, enabling us to assess how effectively we correct biased group-level estimates.

\begin{figure}[ht]
    \centering
    \includegraphics[width=1\linewidth]{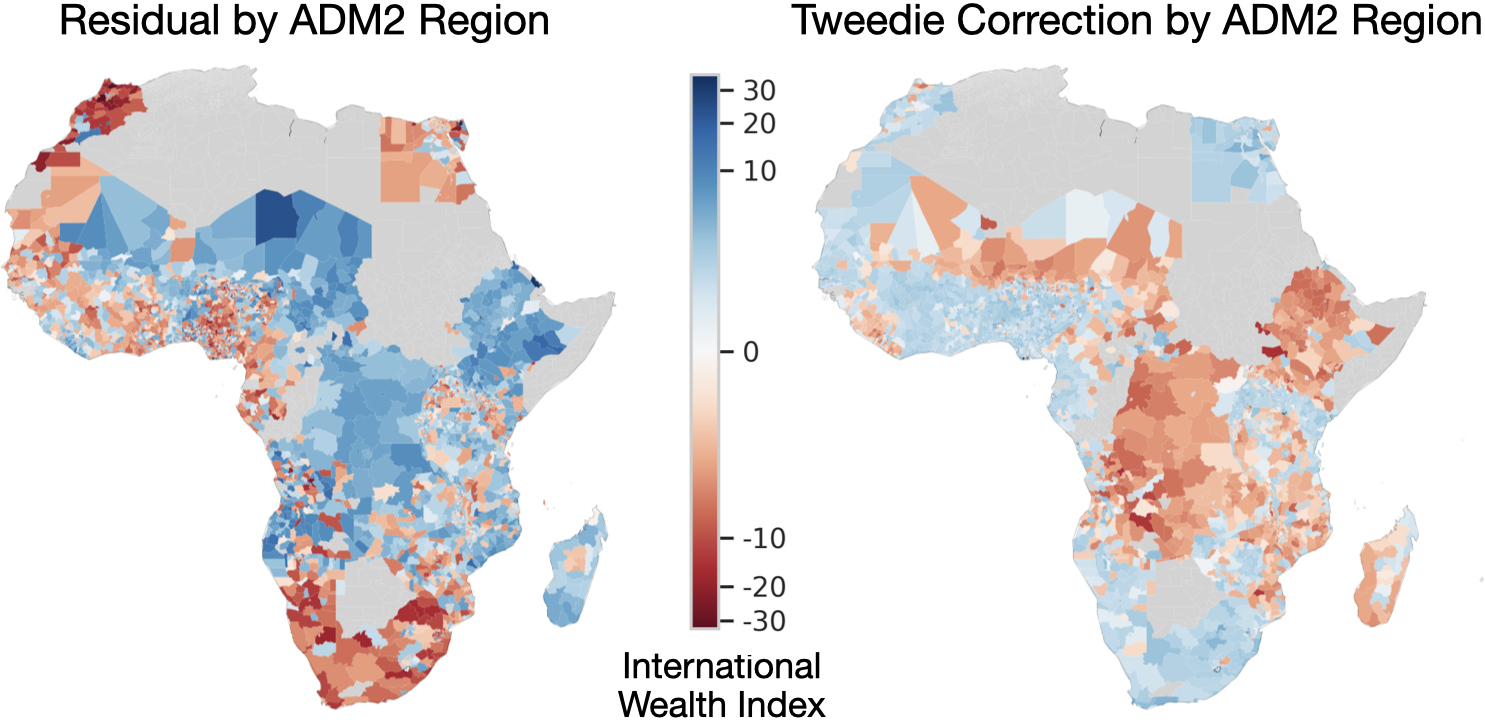}
    \caption{Mean residuals (\textsc{Left}) and adjustments from Tweedie's method (\textsc{Right}) for each ADM2 region in the dataset. Regions that are overestimated in the left plot (blue) are shown to be correctly adjusted down in the right plot (red), and vice versa.
    }
    \label{fig:res_vs_tweedie_maps}
\end{figure}

\begin{figure}[htb]
    \centering

    \begin{subfigure}[t]{0.48\linewidth}
        \centering
        \includegraphics[width=\linewidth]{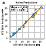}
    \end{subfigure}
    \hfill
    \begin{subfigure}[t]{0.48\linewidth}
        \centering
        \includegraphics[width=\linewidth]{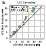}
    \end{subfigure}


    \begin{subfigure}[t]{0.48\linewidth}
        \centering
        \includegraphics[width=\linewidth]{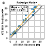}
    \end{subfigure}
    \hfill
    \begin{subfigure}[t]{0.48\linewidth}
        \centering
        \includegraphics[width=\linewidth]{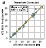}
    \end{subfigure}


    \begin{subfigure}[t]{0.48\linewidth}
        \centering
        \includegraphics[width=\linewidth]{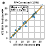}
    \end{subfigure}
    \hfill
    \begin{subfigure}[t]{0.48\linewidth}
        \centering
        \includegraphics[width=\linewidth]{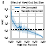}
    \end{subfigure}
    
    \caption{%
    {\it (a–e)} Estimated treatment effects of aid interventions using unadjusted predictions, which display attenuation bias due to shrinkage.
    {\it (f)} PPI outperforms Tweedie’s correction in MAE bias reduction on larger hold-out sets, but requires many ground-truth labels.
    }

  \label{fig:combined_experiment}
\end{figure}

\paragraph{Descriptive Results.} As expected, the MSE of the naive model exhibits shrinkage behavior: it tends to overpredict the wealth of poorer regions and underpredict the wealth of richer regions. As shown in Figure \ref{fig:res_vs_tweedie_maps}, Tweedie's correction accurately compensates for this, not only resulting in less biased estimates (regression slope of 0.83 to 0.90), but also decreases the MAE from 2.67 to 2.39. In fact, if we consider only regions with at least 100 clusters, the Tweedie-corrected estimates exhibit virtually no shrinkage bias, with a slope of 0.99. Appendix C explicates.

\paragraph{Empirical Analysis of Aid Interventions.}
To assess the practical utility of our correction methods in a setting where the true treatment effect is unknown, we construct an evaluation framework based on real-world development projects. Specifically, we leverage the same dataset of geo-referenced interventions funded by the World Bank and the Chinese government as was used in \citep{malik2021banking,Conlin2024}. Each project is tagged with an associated aid sector $s \in \mathcal{S}$ (e.g., \textit{Health}, \textit{Water Supply and Sanitation}, or \textit{Women in Development}), which defines a downstream trial $D_{\text{Trial } s} = \{(\bX_i, \hat{Y}_i, A_{is}): i\in \mathcal{I}_{\textrm{Trial } s}\}$. Here, $A_{is} \in \{0,1\}$ indicates exposure to a sector-$s$ intervention at location $i$, $\bX_i$ are satellite features, and $\hat{Y}_i$ are IWI predictions from the upstream-trained model. Figure \ref{fig:TreatmentSites} depicts the spatial distribution of treatments across the continent. 

Since the treatment effects of these interventions are unknown, we deem this analysis a test of external validity by comparing ML predicted wealth values with observed ones. Our goal is to assess whether bias-corrected predictions can reflect causal estimates similar to those from groundtruth. 

To avoid label leakage and emulate real-world constraints, we perform a two-stage evaluation. We begin by randomly partitioning the available DHS survey data into two disjoint sets: an upstream set for training and calibrating the EO-ML model, and a downstream set used solely for evaluation. The model is trained on the upstream data, without any exposure to the intervention data or knowledge of future evaluation criteria. Shrinkage corrections (e.g., LCC or Tweedie's correction) are also estimated on this upstream set. 

We match project locations to the corresponding ADM2-level administrative regions, roughly equivalent to districts or municipalities. For a given funder-sector pair (e.g., Health projects funded by the World Bank), we identify all ADM2 regions containing at least one relevant intervention site. We then consider villages surveyed in the downstream DHS data within those regions, 3 to 8 years after the intervention, as the treated group. Villages in the same surveys, but outside the treated ADM2 regions, serve as controls.

For each funder-sector combination, we estimate the average treatment effect as the difference in mean outcomes between treated/control villages, for simplicity. We compute this using both observed IWI values and the model-predicted (and correction-adjusted) IWI values. The former provides a benchmark estimate using labeled data, while the latter evaluates the debiasing methods applied to predictions.

\paragraph{Causal Results.} See Figure \ref{fig:combined_experiment} for results. Each point is a causal study. While the $x$-axis shows what would have happened when using the label $Y_i$ in these studies, the $y$-axis shows the corresponding causal effect using $\hat{Y_i}$ with its corresponding adjustment approach. Orange lines check the correlation among these data. A score of one is perfect. The analysis shows that Tweedie's correction (panel \textit{d}) comes close to this desiderata, with a correlation of $0.998$; and it does so with the lowest MAE of $0.57$. Had we used the naive predictions (panel \textit{a}), the correlation would be $0.958$ and the MAE would be $0.69$. Assuming the PPI set up, with 10\% fresh data available, we would have achieved a $0.981$; and it does so with the lowest MAE of $0.73$. The PPI performance is, however, a function of available data in the downstream, and panel \textit{f} shows that with increasing fresh data, the PPI correction will outperform our Tweedie. However, since PPI demands substantial amounts of such data, this method would necessitate costly new surveys, resulting in large expenditures for downstream teams \citep{depieuchon2025benchmarkingdebiasingmethodsllmbased}.

\section{Discussion and Conclusion}\label{s:Discussion}

Our experiments show that {\em post‑hoc}, data–free corrections can eliminate the lion’s share of the shrinkage-induced attenuation that plagues causal analyses based on machine–learning predictions.  In controlled simulations where the true ATE is known, Tweedie’s correction reduces mean–absolute error by an order of magnitude and, crucially, restores calibration: the fitted slope of $\hat{\tau}$ on $\tau$ is indistinguishable from one.  The linear calibration correction also performs nearly as well despite its simplicity, suggesting that much of the bias arises from a first‑order scaling distortion that can be identified on a small held‑out calibration set.  When applied to real DHS imagery, both corrections markedly narrow the gap between predicted and observed country‑level means, and in a quasi‑experimental evaluation of large-scale aid projects, they bring estimated ATE into line with those obtained from ground‑truth survey data. 


That said, bias can remain or be reintroduced when (a) the joint distribution of $(\bX,Y)$ shifts between up and downstream, (b) errors are strongly heteroskedastic or non-Gaussian so that a single $\sigma$ over- or under-corrects in parts of the support, or (c) $\hat\sigma$ is misestimated (e.g., from overfitting late in training). We therefore (i) recommend estimating $\sigma$ on the training split to reduce over-correction, (ii) report block-bootstrap uncertainty for spatial analyses, and (iii) monitor calibration after correction.  \hfill$\square$




\clearpage\newpage

\bibliography{aaai2026}

\appendix









\section{A. Deriving Implications of Tweedie's Formula for Causal Debiasing}\label{app:derive_tweedies}

\input{./tweedie_proofs.tex}

\section{B. Neural Model Details}\label{EOEmpiricalAppendix}

The backbone of each model was a ResNet‑50, pre‐trained via unsupervised contrastive learning (MOCO) on the SSL4EO-L dataset, as distributed in the TorchGeo library \citep{stewart2023ssl4eoldatasetsfoundationmodels}. The images were rescaled following the same procedure as in the SSL4EO-L paper, with added data augmentation in the form of random horizontal and vertical flips. A final dense layer was appended to the ResNet to enable the regression task. 

Training proceeded for up to 33 epochs. During the first three epochs, we froze all pretrained weights and trained only the newly added final layer. Thereafter, we unfroze the entire network and performed end‑to‑end fine‑tuning with early stopping on the validation set (patience = 5 epochs). Optimization used AdamW with weight decay $\left(\ell_2\right.$ regularization coefficient $\left.=1 \times 10^{-4}\right)$, batch size 128 and a cosine learning‐rate schedule with warm restarts; analyses use PyTorch 2.6. When training with the Ratledge objective with AdamW, we \emph{set $\lambda_r=0$} to avoid double-counting $\ell_2$ regularization (weight decay already imposes an $\ell_2$ penalty). 

The only hyperparameter tuning that was performed was to find the value $\lambda_b$ used by the Ratledge loss. This was achieved using a grid search (Figure \ref{fig:ratledge_analysis}).

Quantities $\hat{k}$ and $\hat{m}$ for LCC correction were calculated on the fold's calibration set, values for Tweedie's correction ($\sigma$, the score function) were estimated with the training data. For scores, a Gaussian KDE was fit using Scott's rule. 

Overall, training models for the five cross-validation folds, comprising one standard MSE-based model and eight Ratledge-loss variants, required approximately nine hours of wall-clock time on a single NVIDIA A100 GPU.

\section{C. Additional Empirical Results}\label{EmpiricalAppendix}

For treatment location maps by aid sector and additional figures, see: \texttt{github.com/cjerzak/Figures}. 

\begin{figure}[htb]
\centering
       \includegraphics[width=0.6\linewidth]{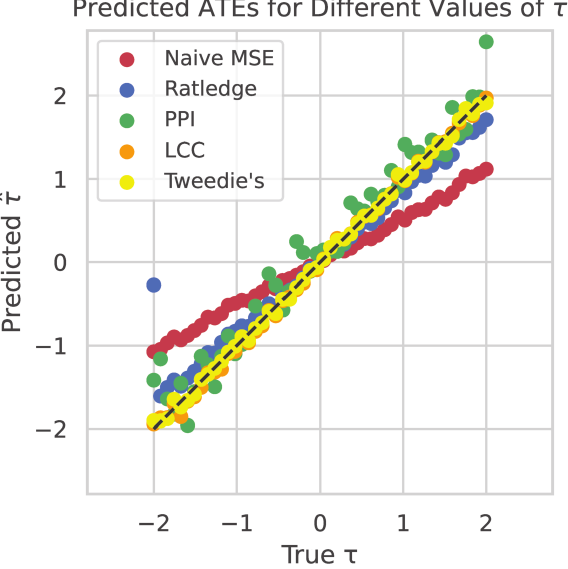}
        \caption{Estimated vs. true ATE across simulated datasets under each method (varying $\tau$); a 45-degree line indicates perfect calibration. 
        }
        \label{fig:simulated_debiased_ate_estimates}        
      \end{figure}

\begin{figure}[htb]
\centering
       \includegraphics[width=1\linewidth]{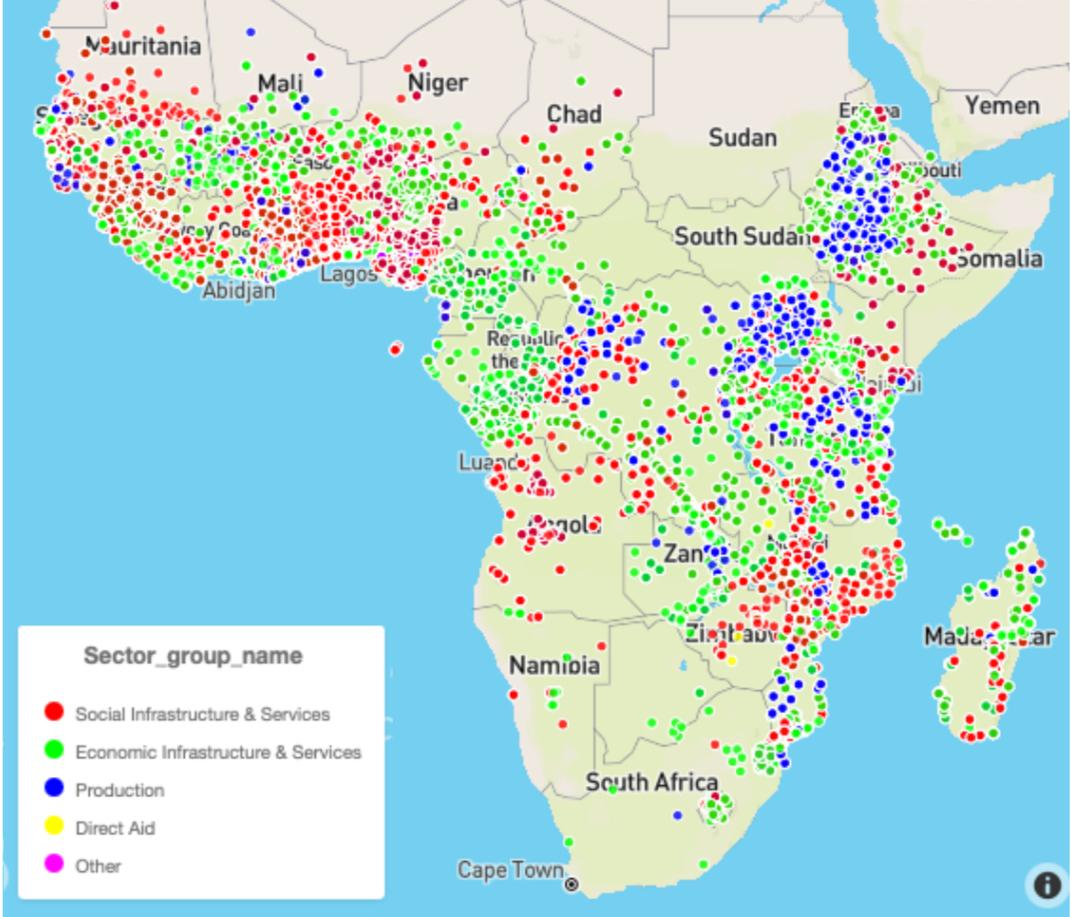}
       \caption{
         Treatment sites.
       }\label{fig:TreatmentSites}
      \end{figure} 

\begin{figure}[htb]
    \centering
    \includegraphics[width=1\linewidth]{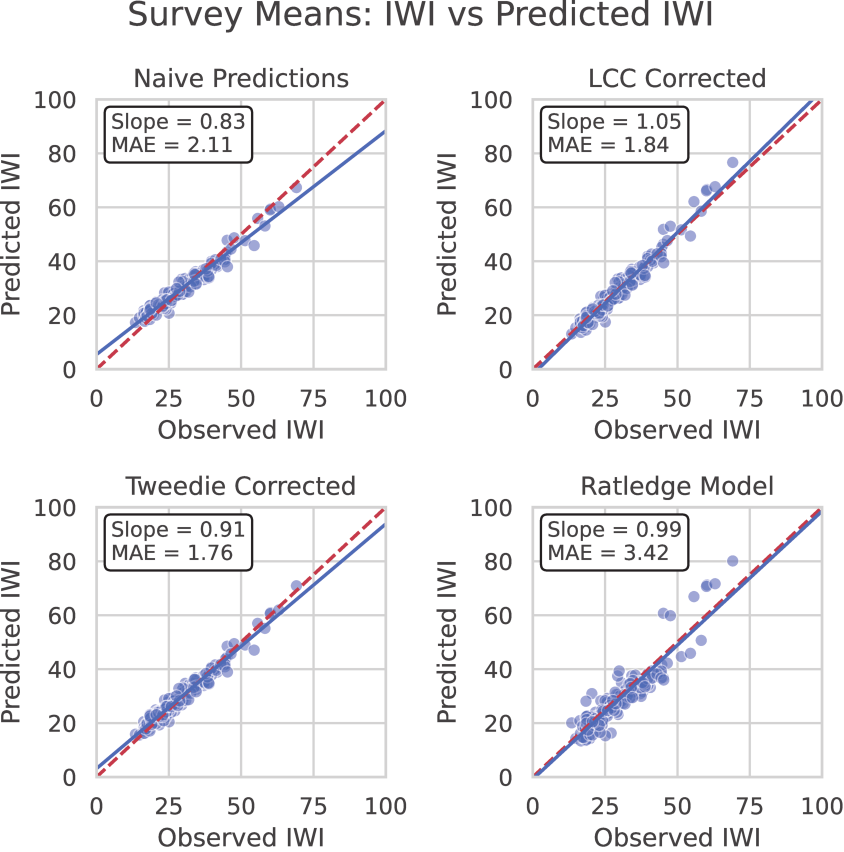}
    \caption{
    Performance comparing MAE and calibration slope of Tweedie and Ratledge adjustment, seen via scatterplot of observed versus predicted mean IWI for each survey in the test set. The standard MSE-trained model produces biased estimates, overestimating wealth in poorer countries and underestimating it in richer ones. Applying Tweedie's or LCC adjustments significantly mitigates this shrinkage bias. Training a model with Ratledge loss also mitigates shrinkage bias in this context, but results in higher variance than with Tweedie adjustment (seen by lower MAE for Tweedie).
    }
\label{fig:survey_estimates}
\end{figure}

\begin{figure}[htb]
    \centering
    \includegraphics[width=1\linewidth]{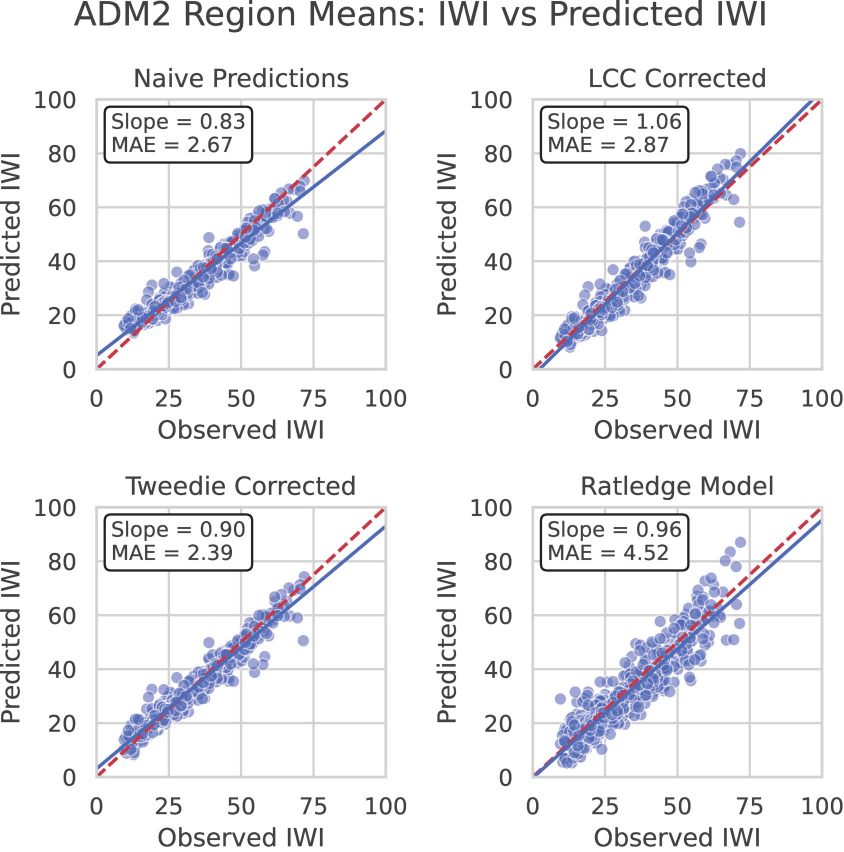}
    \caption{
    Scatterplot of the observed versus predicted mean IWI for each Admin 2 region in the test set, similar to \ref{fig:survey_estimates}. The standard MSE-trained model produces biased estimates, overestimating wealth in poorer countries and underestimating it in richer ones. Applying Tweedie's or LCC adjustments significantly mitigates this shrinkage bias. Training a model with Ratledge loss also mitigates the shrinkage bias, but has a relatively high error in terms of MAE.
    }
\label{fig:adm2_estimates}
\end{figure}

\begin{figure}[htb]
    \centering
    \includegraphics[width=1\linewidth]{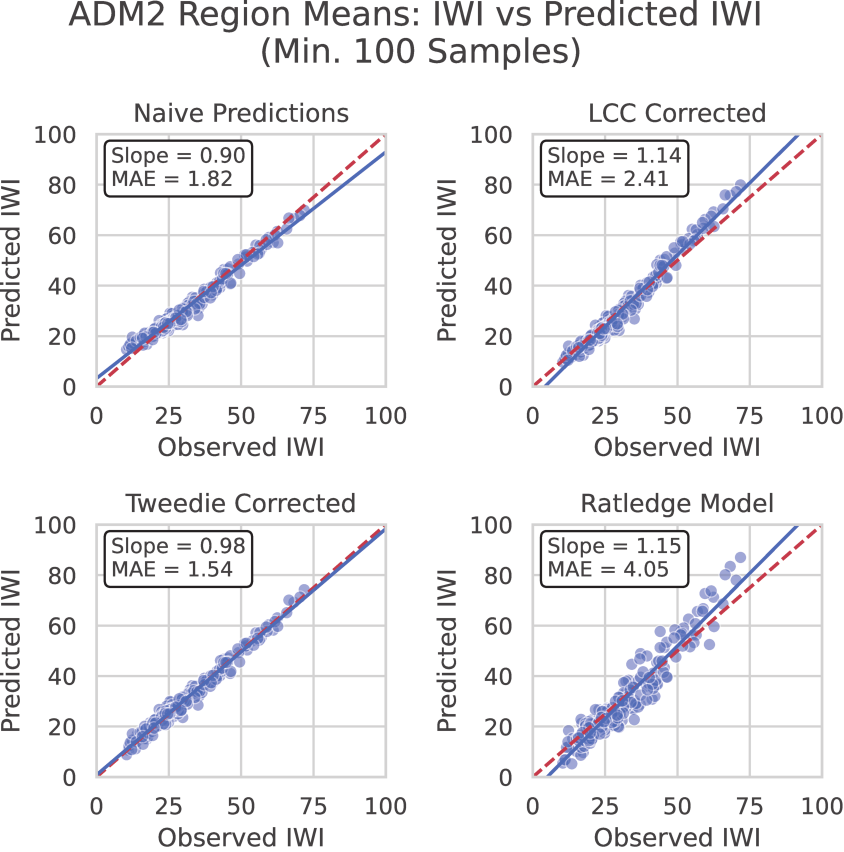}
    \caption{
    Same results as for \ref{fig:adm2_estimates}, but only including Admin 2 regions with more than 100 surveyed clusters. The Tweedie adjusted model is now practically unbiased (in regards to shrinkage), suggesting that the correction has larger issues with smaller samples.
    }
\label{fig:adm2_estimates_min_100}
\end{figure}

\begin{figure}[htb]
    \centering
    \includegraphics[width=1\linewidth]{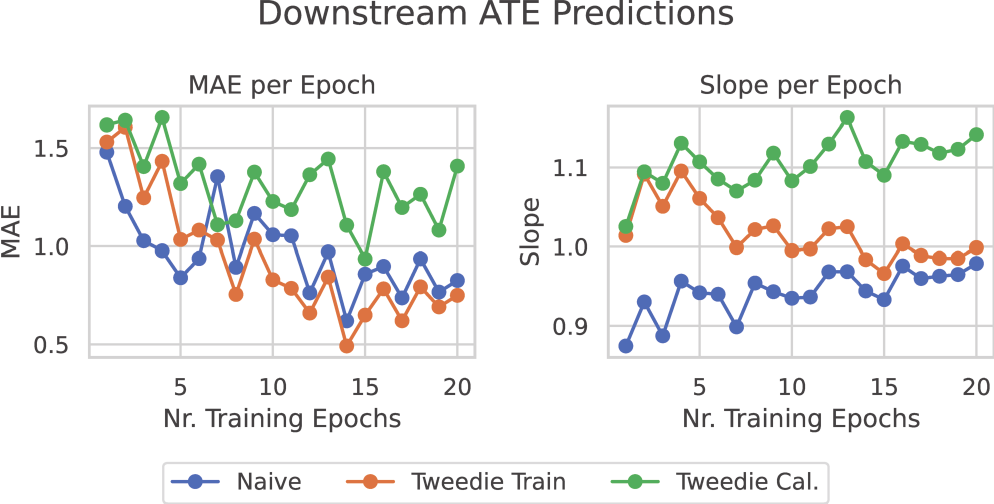}
    \caption{
Performance on a downstream causal inference task as a function of upstream model training epochs. Early training (e.g., after one epoch) yields poor model accuracy (high MAE) but minimal overfitting, allowing Tweedie's correction to produce unbiased estimates using variance from either the calibration set ($\sigma_{\text{Cal.}}$) or training set ($\sigma_{\text{Train}}$). As epochs increase, model accuracy improves, but growing overfitting causes over-correction when using $\sigma_{\text{Cal.}}$, increasing bias, while $\sigma_{\text{Train}}$ maintains near-unbiased results.
}
    \label{fig:EpochFig}
\end{figure}

\subsection{Sensitivity to Calibration Data}

The DHS displaces survey coordinates ($\geq$ 3 km) to protect privacy, so we already address a built‑in corruption source, which shows improvement. Generally, a weak method or noisy data would make debiasing more critical, as uncertainty increases shrinkage. To show this, we ran an additional sensitivity analysis on the causal aid task (cf. Fig 6). We trained our model on the same dataset but shuffled a share of the labels in the train+cal to add noise, only increasing Tweedie’s relative advantage. Results are below in Tables \ref{tab:BadNaive} and \ref{tab:BadTweedie}.

\begin{table}[htb]
\centering\small
\begin{tabular}{lll}
\hline
\textbf{Shuffled frac.} & \textbf{Slope} & \textbf{MAE} \\ \hline
0              & 0.96  & 0.69 \\
0.05           & 0.94  & 0.67 \\
0.15           & 0.77  & 2.36 \\
0.30           & 0.64  & 3.91 \\ \hline
\end{tabular}
\caption{Performance of the naive estimator under varying levels of label shuffling in training and calibration data.}\label{tab:BadNaive}
\end{table}

\begin{table}[htb]
\centering\small
\begin{tabular}{lll}
\hline
\textbf{Shuffled frac.} & \textbf{Slope} & \textbf{MAE} \\ \hline
0                       & 1.00           & 0.57         \\
0.05                    & 1.04           & 0.69         \\
0.15                    & 0.93           & 1.14         \\
0.30                    & 1.05           & 1.32         \\ \hline
\end{tabular}
\caption{Performance of Tweedie's estimator under varying levels of label shuffling in training and calibration data.}\label{tab:BadTweedie}
\end{table}

\noindent\emph{Variance source for Tweedie.} Figure \ref{fig:EpochFig} shows that estimating $\sigma$ on the training split (rather than the calibration split) mitigates over-correction at later epochs; our package therefore exposes both options, defaulting to $\sigma_{\text{Train}}$. We also present in \ref{tab:SensitivitySize} the results of an analysis probing the sensitivity of correction performance to cal set size (mean $\pm$ sd across 20 random subsamples). 

\begin{table}[htb]
\centering
\scriptsize
\begin{tabular}{lllll}
\textbf{n cal.} & \textbf{LCC slope} & \textbf{LCC MAE} & \textbf{Tweedie's slope} & \textbf{Tweedie's MAE} \\ \hline
50              & 1.23 ± 0.05        & 3.17 ± 0.57      & 1.00 ± 0.01              & 0.63 ± 0.04            \\
500             & 1.22 ± 0.01        & 2.99 ± 0.17      & 1.00 ± 0.00              & 0.60 ± 0.01            \\
5000            & 1.22 ± 0.00        & 2.99 ± 0.04      & 1.01 ± 0.00              & 0.59 ± 0.01            \\ \hline
\end{tabular}
\caption{Sensitivity of LCC and Tweedie's estimators to calibration set size (mean $\pm$ sd across 20 random subsamples).}\label{tab:SensitivitySize}
\end{table}




\subsection*{When Is Correction Warranted?}

To detect and mitigate attenuation bias in model predictions, begin by evaluating them against true labels on a held-out calibration dataset to assess whether predictions systematically shrink toward the mean, as indicated by a regression slope less than one with a confidence interval excluding one. If such bias is confirmed, apply an appropriate correction: for cases where the relationship appears broadly linear, use a simple linear calibration adjustment to rescale the predictions; for more complex scenarios involving nonlinearity or pronounced shrinkage in extreme values, employ a local adjustment method like Tweedie's correction, leveraging density estimates and noise levels derived from upstream training. Finally, verify the correction's effectiveness by re-evaluating the calibration slope on the adjusted predictions to ensure its confidence interval includes one, and for spatially correlated data, incorporate block bootstrapping to account for error dependencies in uncertainty estimates.

\begin{table*}[htb]
\centering\small
\begin{tabular}{l p{4cm} p{3.5cm} p{4cm}}
\toprule
\textbf{Method} & \textbf{Data Needs} & \textbf{Training Adjustments} & \textbf{Other Assumptions} \\
\midrule
{\it (i) PPI \citep{angelopoulos2023prediction}} & Fresh ground-truth labeled data at the downstream causal inference stage (new treatment-predictor-outcome tuples). & None; treats model as black-box and applies rectifier term. & Predictive model is fixed (not retrained); valid statistical coverage in large samples; independence assumptions for inference. \\
\addlinespace
{\it (ii) \citet{ratledge_using_2022}} & None additional; uses existing upstream training data. & Yes; modifies loss function to include quintile-specific bias penalty ($\mathcal{L}=\mathrm{MSE}+\lambda_r\|\theta\|_2^2+\lambda_b\hat{E}_b$; if using AdamW, set $\lambda_r=0$).
Requires retraining. & Bias reducible by penalizing group-level discrepancies; assumes quintile-based error structure. \\
\addlinespace
{\it (iii) DSL \citep{dsl_original}} & Gold-standard labeled data on a subset of units (no new data collection); known or estimated sampling probabilities depending on surrogates and covariates. & None; post-hoc construction of bias-corrected pseudo-outcomes via cross-fitting and doubly-robust estimation. & Known sampling design with probabilities bounded away from zero; double robustness (consistent if either outcome model or sampling model is correctly specified). \\
\addlinespace
{\it (iv) LCC (this work)} & Held-out calibration data from upstream training phase (no new data collection). & None; post-hoc linear transformation on predictions. & Linear shrinkage model ($\hat{Y}_i = k Y_i + m$ with $0 < k \leq 1$). \\
\addlinespace
{\it (v) Tweedie's correction (this work)} & Data from upstream for estimating $\sigma^2$ and score function (no new data collection). & None; post-hoc adjustment using empirical Bayes and density scores. & Gaussian homoskedastic noise ($\hat{Y}_i = Y_i - \varepsilon_i$, $\varepsilon_i \sim \mathcal{N}(0, \sigma^2)$ independent of $Y_i$); $\mathbb{E}[\varepsilon_i \mid A_i] = 0$. \\
\bottomrule
\label{table:Debiasing}
\end{tabular}
\caption{Characterization of Debiasing Methods for Causal Inference Using ML Predictions}
\label{tab:debiasing-methods}
\end{table*}

\begin{table*}[htb]
    \centering\scriptsize
    \begin{tabular}{lllccccc}
\toprule
Funder &  Sector &                                     Aid Sector Name &   $\widehat{\textrm{ATE}}^{Y}$ & $\widehat{\textrm{ATE}}^{\hat{Y}}$ & $\widehat{\textrm{ATE}}^{\hat{Y}}$, Tweedie &  \# Treated & \# Control \\
\midrule
    CH &             110 &                                          Education &         13.02 &                13.01 &                   13.36 &                   1549 &                  22381 \\
    WB &             110 &                                          Education &          7.33 &                 7.77 &                    7.79 &                   6285 &                  24262 \\
    WB &             120 &                                             Health &          3.75 &                 4.05 &                    4.11 &                   6965 &                  24411 \\
    CH &             120 &                                             Health &         19.71 &                19.38 &                   19.86 &                   3333 &                  32198 \\
    CH &             130 & Population Policies/Programmes &         11.30 &                 9.06 &                    9.42 &                     60 &                    335 \\
    WB &             140 &                        Water Supply and Sanitation &          8.04 &                 7.77 &                    7.86 &                  10470 &                  27888 \\
    CH &             140 &                        Water Supply and Sanitation &         19.08 &                17.32 &                   17.69 &                    676 &                   7883 \\
    CH &             150 &                       Government and Civil Society &         21.55 &                20.88 &                   21.31 &                   1860 &                  22983 \\
    WB &             150 &                       Government and Civil Society &          2.18 &                 2.47 &                    2.33 &                  16027 &                  25739 \\
    WB &             160 &           Other Social Infrastructure and Services &          5.48 &                 5.28 &                    5.65 &                   9178 &                  25226 \\
    CH &             160 &           Other Social Infrastructure and Services &         20.15 &                19.40 &                   19.75 &                    843 &                  13557 \\
    CH &             210 &                              Transport and Storage &         20.16 &                18.79 &                   19.19 &                   1509 &                  14122 \\
    WB &             210 &                              Transport and Storage &          3.81 &                 3.99 &                    3.66 &                  12550 &                  28163 \\
    WB &             220 &                                     Communications &          0.50 &                 1.60 &                    0.84 &                   1537 &                  11327 \\
    CH &             220 &                                     Communications &         12.42 &                12.22 &                   12.36 &                    997 &                   6371 \\
    WB &             230 &                       Energy Generation and Supply &          7.16 &                 7.23 &                    7.35 &                   6683 &                  25702 \\
    CH &             230 &                       Energy Generation and Supply &         13.00 &                13.07 &                   12.97 &                    357 &                   7958 \\
    WB &             240 &                     Banking and Financial Services &          8.14 &                 7.63 &                    8.30 &                   2363 &                  14997 \\
    WB &             310 &                  Agriculture, Forestry and Fishing &          3.21 &                 3.09 &                    3.10 &                   8907 &                  25187 \\
    CH &             310 &                  Agriculture, Forestry and Fishing &          4.49 &                 4.35 &                    3.78 &                    949 &                  10400 \\
    WB &             320 &                     Industry, Mining, Construction &          2.80 &                 2.86 &                    2.95 &                   5183 &                  23180 \\
    CH &             320 &                     Industry, Mining, Construction &         21.31 &                23.75 &                   24.21 &                     74 &                   2041 \\
    WB &             330 &                                  Trade and Tourism &          2.63 &                 4.07 &                    3.46 &                    840 &                   7852 \\
    CH &             330 &                                  Trade and Tourism &         19.10 &                20.12 &                   20.48 &                     52 &                    487 \\
    WB &             410 &                     General Environment Protection &         13.06 &                14.03 &                   13.94 &                   1284 &                  11273 \\
    CH &             420 &                               Women in Development &         18.95 &                18.53 &                   18.89 &                     81 &                   3238 \\
    CH &             430 &                                  Other Multisector &         26.74 &                25.06 &                   25.93 &                    225 &                   3755 \\
    CH &             520 &    Developmental Food Aid/Food Security &         18.20 &                17.89 &                   18.92 &                    170 &                   2163 \\
    CH &             700 &                                 Emergency Response &         11.24 &                10.54 &                   10.74 &                    258 &                   5090 \\
\bottomrule
\end{tabular}
\caption{
Sector effect estimates, where the outcome scale is IWI units. Here, $\widehat{\textrm{ATE}}^{Y}$ refers to estimates of treatment effects using difference-in-means with DHS outcomes (considered to be ground truth). $\widehat{\textrm{ATE}}^{\hat{Y}}$ refers to estimates with predicted outcomes. 
}
    \label{tab:my_label}
\end{table*}

  \begin{table*}[htb]
      \centering\scriptsize
      \begin{tabular}{lrrlrr}
  \toprule
          &             & \multicolumn{2}{c}{---IWI---}                &              &          \\
   Country & \# Clusters & Mean                    &      SD     & Survey Years & Urban \% \\
  \midrule
  
         Angola &   969 & 32.68 & 24.06 &                                   06-07, 11, 15-16 & 50.2 \\
          Benin &  1567 & 28.82 & 13.31 &                               96, 01, 11-12, 17-18 & 43.7 \\
   Burkina Faso &  2232 & 26.52 & 12.18 &                   93, 98-99, 03, 10, 14, 17-18, 21 & 32.8 \\
        Burundi &  1128 & 19.01 & 10.81 &                                      10, 12, 16-17 & 19.2 \\
         C.A.F. &    90 & 13.52 &  4.57 &                                              94-95 & 30.0 \\
       Cameroon &  1920 & 35.95 & 17.23 &                                 91, 04, 11, 18, 22 & 53.6 \\
           Chad &   624 & 16.25 & 10.74 &                                              14-15 & 26.1 \\
        Comoros &   242 & 38.37 & 12.38 &                                                 12 & 43.8 \\
   Côte d’Ivoire &  1113 & 37.84 & 14.19 &                               94, 98-99, 11-12, 21 & 52.3 \\
         D.R.C. &   783 & 18.34 & 14.04 &                                          07, 13-14 & 35.9 \\
        Egypt &  7637 & 59.39 & 11.79 &                         92, 95, 00, 03, 05, 08, 14 & 48.2 \\
     Eswatini &   270 & 38.00 & 13.50 &                                              06-07 & 40.4 \\
     Ethiopia &  2532 & 17.78 & 15.29 &                                 00, 05, 11, 16, 19 & 29.2 \\
\midrule
        Total & 68619 & 32.59 & 19.11 &                                                  - & 37.8 \\
\bottomrule
   \end{tabular}
        \caption{Overview of data: DHS surveys on health and living conditions.}
    \label{tab:my_label2}
  \end{table*}


\section{D. Simulation Design}\label{SimulationAppendix}
In this DAG, $C_i$ is a confounder that influences both the treatment assignment $A_{is}$ and the outcome $Y_i$. Our simulation samples it from a standard Normal distribution, $C_i \sim \mathcal{N}(0, 1)$. The treatment $A_{is}$ (we suppress the index $s$ henceforth) is assigned probabilistically based on $C_i$ using a sigmoid function:
\begin{equation}
    A_{i} \sim \text{Bernoulli}\left( \frac{1}{1 + \exp(-C_i)} \right)
    \label{eq:treatment_prob}
\end{equation}
The sigmoid maps the confounder ($C_i$) to a probability in that governs Bernoulli treatment assignment. The outcome, $Y_i$, is then generated as:
\begin{equation*}
Y_i \sim \mathcal{N}(\tau A_i + C_i, \sigma_Y^2)
\end{equation*}
This formulation implies that higher values of $C_i$ both increase the likelihood of treatment and the expected outcome.

To emulate a real-world scenario in which outcomes are not directly observed but inferred from complex data (e.g., satellite imagery), we construct a latent representation $\bX_i = g(Y_i)$. Here, $g$ is a randomly initialized three-layer neural network with ReLU activations. The network takes scalar $Y_i$ as input, processes it through hidden layers of dimension 50, and outputs a 100-dimensional embedding $\bX_i$.

A separate model, $f$, is then trained to predict the outcome, $Y$, from the embedding, $\bX_i$. The resulting predictions $\hat{Y}_i = f(\bX_i)$ are used to estimate the treatment effect $\tau$ via Inverse Probability of Treatment Weighting (IPTW), with the treatment probability given as in Eq. \ref{eq:treatment_prob}. To compare different debiasing approaches, IPTW estimates were computed for various predicted values.

We run an experiment to simulate a full prediction-to-causal-inference pipeline using synthetic data. For a given value of $\tau$, we generate a training population according to the predefined causal DAG. To construct embeddings $\bX_i$, we initialize a neural network $g$ with random, frozen weights. This network maps the scalar outcome $Y_i$ to a 100-dimensional embedding: $\bX_i = g(Y_i)$. This setup induces a non-linear and noisy relationship between $Y_i$ and $\bX_i$, mimicking real-world scenarios where predictive features encode partial and indirect information about outcomes.

We then train a separate neural network $f$ to predict $Y_i$ from $\bX_i$ using the standard mean squared error loss, producing predicted values $\hat{Y}_i = f(\bX_i)$. A test population is generated using the same DAG, parameter settings, and embedding function, $g$. From this test population, we compute multiple different estimates of the treatment effect, as described below, using the trained model outputs with corrections. To evaluate the robustness of our findings to the hyperparameter choice of $\tau$, we repeat this entire process across 51 evenly spaced values of $\tau$ in the range $[-2, 2]$. On average, the predictions, $\hat{Y}_i$, of model $f$ get a mean $R^2$ value of about $0.5$ when used to predict $Y_i$. 

First, we compute the {\sc Sample ATE} using inverse probability weighting on the true labels. As this is based on finite samples, it provides a noisy, but approximately unbiased, estimate of $\tau$. Next, we compute the {\sc Predicted ATE} using the predicted outcomes $\hat{Y}_i$ instead of $Y_i$. As expected, this estimate exhibits shrinkage bias, systematically pulling $\hat{\tau}$ toward zero. Finally, we apply the four debiasing methods mentioned above to the predicted outcomes to produce corrected estimates.

\section{E. World Bank and China Empirical Cases}\label{SimulationAppendix}

\begin{figure}[htb]
    \centering
    \includegraphics[width=1\linewidth]{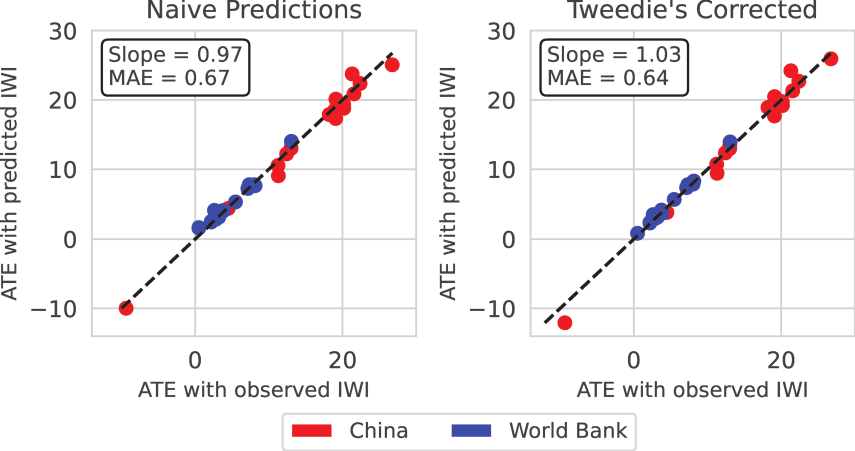}
    \caption{
      Scatterplots comparing ATE estimated using predicted IWI versus observed IWI for aid interventions funded by China (red points) and the World Bank (blue points). Tweedie's correction slightly improves the overall MAE.
    }
    \label{fig:china_vs_wb}
\end{figure}

For robustness, we repeat this experiment, using urban/rural, country, and survey year as confounders (Figure \ref{fig:china_vs_wb}). The results show that this approach breaks the bias in terms of treatment effect. For larger studies, with more than 200 treated points, it also lowers the total error rate. We hypothesize that the reason the overall error rate does not decrease for small trials is due to the added variance introduced when calculating the Tweedie correction. The correction is a mean of the sample scores; it decreases with larger sample sizes.

\begin{figure}[htb]
    \centering
    \includegraphics[width=1\linewidth]{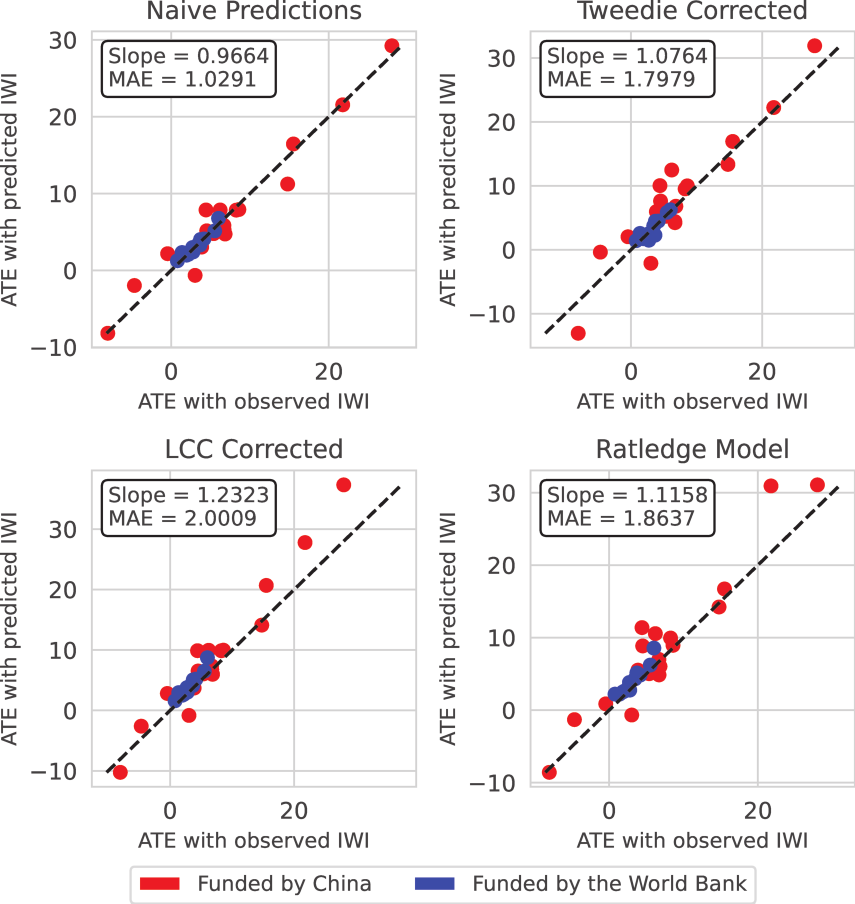}
    \caption{
      Scatterplots comparing ATE estimates by funder when country, year and urban/rural status are used as confounders. Unlike in Figure \ref{fig:china_vs_wb}, the samples were weighted using IPTW to account for these confounders.
    }
    \label{fig:iwi-plot}
\end{figure}

\section{F. Implementing Strategies with \texttt{unshrink}}\label{PackageAppendix}

For more information, see: 
\begin{itemize}
    \item[] \hspace{-0.53cm} {\scriptsize \texttt{github.com/AIandGlobalDevelopmentLab/unshrink} }
\end{itemize}
For all-Africa IWI wealth map predictions with associated score and calibration parameters (e.g., $\sigma$) needed for adjustment:
\begin{itemize}
    \item[] \hspace{-0.50cm}{\scriptsize \texttt{aidevlab.org/AfricaIWIv1} }
\end{itemize}
For the code appendix, please see:
\begin{itemize}
    \item[] \hspace{-0.50cm}{\scriptsize\texttt{github.com/AIandGlobalDevelopmentLab/OneMapManyTrials} }
    \end{itemize}

    \begin{figure}[htb]
    \centering
    \includegraphics[width=1\linewidth]{Figures/ratledge_analysis_compressed_pdfa.pdf}
    \caption{Results of the hyperparameter tuning for finding the best suited $\lambda_b$ for the Ratledge loss. Based of these results, we opted to use $\lambda_b=15$ as it had the least shrinkage bias $(b \approx 1)$.}
    \label{fig:ratledge_analysis}
\end{figure}

\end{document}

%% file: tweedie_main.tex

\subsection{Tweedie's Correction}
\label{s:Tweedie}

To go beyond the linearity assumption in LCC in a way that respects the structure of ML predictions, we now turn to Tweedie’s identity, which provides a local, density–score–based way to reverse shrinkage without assuming global linearity of the measurement error process.

A first instinct might be to formalize shrinkage in a way that generalizes beyond the linearity assumption of LCC by using a measurement-error model, in which the prediction is a noisy version of the truth:
\[
\hat{Y}_i = Y_i + \varepsilon_i,\; \;\; \varepsilon_i \sim \mathcal{N}(0, \sigma^2) 
\]
However, in ML predictions scenarios, models approximate \(\mathbb{E}[Y_i \mid \bX_i]\) for covariates $\bX_i$ and thus predictions tend to exhibit systematically lower variance than targets due to shrinkage toward the mean (the opposite of what is implied in the baseline classical case, where $\textrm{Var}(\hat{Y}_i) > \textrm{Var}(Y_i)$). Indeed, upstream ML models are trained to minimize expected loss over $\bX_i$ \citep{shalev2014understanding}:
\[
Y_i = \hat{Y}_i + \epsilon_i, \quad \hat{Y}_i=f_{\theta}(\mathbf{X}_i)
\]
where $\theta$ are neural parameters optimized to reduce MSE. Any residual noise $Y_i - \hat{Y}_i$ thus arises \emph{after} the prediction, approximately independent of $\hat{Y}_i$ itself in distribution (as the model would otherwise exploit correlations to reduce loss). This reasoning therefore inverts the classical view, motivating adoption of the Berkson error model \citep{carroll2006measurement,heid2004two}:
\[
Y_i = \hat{Y}_i + \varepsilon_i,\; \;\; \varepsilon_i \sim \mathcal{N}(0, \sigma^2) 
\perp \hat{Y}_i,
\]
as a more natural framing for debiasing ML surrogates.

\vspace{0.35\baselineskip}
\paragraph{Adapting Tweedie's Formula.}
To reverse shrinkage under this model without fresh labels, we propose an approach using \emph{Tweedie's formula}, introduced by \citet{robbins1956empirical} and commonly encountered in Gaussian diffusion models:
\begin{itemize}
\item[] ---\textit{Tweedie's Formula:}---
\item[] For any random variable, $Z$, with $Z=\omega+\varepsilon$, 
\\ where $\varepsilon\sim\mathcal N(0,\sigma^{2})$,
\item[] \;\; $\E[\omega\mid Z=z]=z+\sigma^{2}\,\frac{d}{dz}\log p_Z(z)$\;\; \citep{efron2011tweedie}.
\end{itemize}
In its native form, Tweedie's formula under the Berkson model gives us an expression for $\mathbb{E}[ \hat{Y}_i \mid Y_i = y]$, i.e., recovering model predictions under attenuation bias from true labels. Our aim is to derive a novel debiasing method that works in the opposite direction, i.e., recovering estimates of the true labels without attenuation bias in downstream applications. We achieve this goal by introducing a new-label-free corrected pseudo-outcome, \(\widetilde Y_i\), whose conditional mean recovers the truth, \(\mathbb{E}[\widetilde Y_i\mid Y_i]=Y_i\). To obtain this end, we need to deploy the score-swapping relation:
$$
\frac{d}{dy} \log p_Y(y) = \mathbb{E} \left[ \frac{d}{d\hat{y}} \log p_{\hat{Y}}(\hat{Y}_i) \mid Y_i = y \right],
$$
as derived in Appendix I, to the Tweedie identity:
\begin{align*}
\small
    \mathbb{E}\left[ \hat{Y}_i \mid Y_i = y \right] &= y + \sigma^2 \frac{d}{dy} \log p_Y(y) \\
    &= y + \sigma^2 \, \mathbb{E}\left[\frac{d}{d\hat{y}} \log p_{\hat{Y}}(\hat{Y}_i) \mid Y_i = y \right] \\
    \Longrightarrow y &= \mathbb{E}\left[\hat{Y}_i - \sigma^2  \frac{d}{d\hat{y}} \log p_{\hat{Y}}(\hat{Y}_i) \mid Y_i = y \right],
\end{align*}
where the move from the second to the third line follows from linearity of expectations. 

Intuitively, subtraction in the correction arises from inverting the forward application of Tweedie's formula under the Berkson model. Unconditionally, $\E[\hat{Y}_i]=\E[Y_i]$, so predictions are centered around the true values on average (as ensured by MSE minimization during training). Conditionally, however, $\E[\hat{Y}_i \mid Y_i=y]=y + \sigma^2 \frac{d}{dy} \log p_Y(y)$, where the score term attenuates extreme predictions toward the population mean---for instance, when $y$ exceeds the mode of $p_Y$, the score is negative and pulls the conditional expectation below $y$ and towards the mode. In other words, while the overall distribution of predictions matches that of the truths in expectation, a given prediction (absent adjustment) shrinks toward the mean conditional on the realized true value. The inversion thus requires subtracting the score---now estimated via the swapped relation on $p_{\hat{Y}}$---to reverse this attenuation and expand the tails.

This discussion then yields the Tweedie corrected pseudo-outcome under the Berkson model:
\begin{equation}
\widetilde Y_i = \hat Y_i - \sigma^2 \, \frac{d}{d\hat y} \log p_{\hat Y}(\hat Y_i).
\label{eq:po-berkson}
\end{equation}
The score term expands both tails: when the score is positive (where we evaluate the score rising to the left of a local mode), it pushes \(\hat Y_i\) downward; when the score is negative (score evaluated to the right of a local mode) it pushes \(\hat Y_i\) upward; around a mode, where the score is near zero, adjustment is negligible. In practice, $\sigma$ and the score function will need to be estimated on data: $\sigma$ is estimated from residuals; the score function with, e.g., a KDE on $\{\hat{Y}_i\}_{i=1}^n$. 

\vspace{0.35\baselineskip}
\begin{proposition}[Tweedie calibration identity]
\label{prop:berkson_calibration_identity}
The pseudo-outcome satisfies: 
$\mathbb{E}\!\left[\widetilde Y_i \,\middle|\, Y_i\right] \;=\; Y_i.$
\end{proposition}
\emph{Proof sketch.}
By Tweedie’s identity, $\E[\hat Y_i\mid Y_i=y] = y + \sigma^2\,\frac{d}{d y}\log p_{Y}(y)$.
By score‑swapping (Appendix I), $\frac{d}{d y}\log p_Y(y) = \E[ \frac{d}{d \hat Y} \log p_{\hat{Y}}(\hat Y_i)\mid Y_i=y]$.
Take conditional expectations of $\widetilde Y_i$ given $Y_i=y$ to obtain $y$.

\noindent \textbf{Remark}. This is the key statement giving rise to well-calibrated downstream applications of the ``one-map'' ML prediction. Note that the variance of $\tilde{Y}_i$ will almost certainly be larger than the variance of $\hat{Y}_i$, so for purely predictive purposes, the overall MSE of pseudo-outcomes may also be larger that of $\hat{Y}_i$; however, downstream teams will for causal inference purposes be concerned with well-calibrated group conditional means---discussed later. 

\begin{proposition}[Local Reduction to LCC]
\label{prop:local_LCC}
If $p_{\hat Y}$ is locally Gaussian with mean $\mu$ and variance $\gamma^2$,
then the score is $-(\hat y-\mu)/\gamma^2$ and the pseudo‑outcome is the affine map:
\begin{align*} 
\widetilde Y_i
& \;=\; \hat Y_i-\sigma^2\left(-\frac{\hat Y_i-\mu}{\gamma^2}\right) 
= 
\mu + \Bigl(1+\tfrac{\sigma^2}{\gamma^2}\Bigr)(\hat Y_i-\mu),
\end{align*}
i.e., a linear calibration with slope $(1+\sigma^2/\gamma^2)$ increasing in the ratio of measurement error variance to local signal variance that expands extremes away from $\mu$.
\end{proposition}
\emph{Proof.} Plug the Gaussian score into $\widetilde Y_i$.

\vspace{0.35\baselineskip}
\paragraph{Further Relations to LCC.} When the score term is negligible (e.g., flat density), \(\widetilde Y_i \approx \hat Y_i\), reverting to uncorrected predictions. While LCC provides a single global scaling adjustment for errors that operates in a uniform manner, Tweedie adjustment here can handle more complex distributional patterns, rescaling depending on the position of predictions relative to local (possibly multiple) modes. Thus, LCC captures a single regression-to-the-mean dynamic, while Tweedie adjustment captures a broader range of error processes, adjusting for regression to local modes.

\begin{proposition}[Unbiasedness of Tweedie means]
Under the assumptions of Proposition~\ref{prop:berkson_calibration_identity}, for any partitioning variable $A_i$ 
\label{prop:berkson_ate}
such that $\hat{Y}_i \perp A_i \;|\; Y_i$ (which implies $\tilde{Y}_i \perp A_i \; | \; Y_i$ [because $\tilde{Y}_i$ is a deterministic function of $\hat{Y}_i$]):
\[
\E[\widetilde Y_i\mid A_i=a] \;=\; \E[Y_i\mid A_i=a].
\]
Consequently, the difference‑in‑means estimator
\(
\hat\tau:=\hat{\E}[\widetilde Y_i\mid A_i=1]-\hat{\E}[\widetilde Y_i\mid A_i=0]
\)
satisfies $\E[\hat\tau]=\E[Y_i\mid A_i=1]-\E[Y_i\mid A_i=0]$.
Under random assignment, $\E[\hat\tau]=\tau$.
\end{proposition}
\emph{Proof sketch.}
Iterated expectations using Proposition~\ref{prop:berkson_calibration_identity} yield
$\E[\widetilde Y_i\mid A_i=a]
=\E\{\E[\widetilde Y_i\mid Y_i, A_i=a]\mid A_i=a\}
=\E[Y_i\mid A_i=a]$. Independence of $A_i$ and $\epsilon_i$ is key here, since it implies that there is no additional relationship between $\tilde{Y}_i$ and $Y_i$ once $Y_i$ is known, so $\mathbb{E}[\tilde{Y}_i|Y_i]=\mathbb{E}[\tilde{Y}_i|Y_i,A]$. Randomization identifies $\E[Y_i\mid A_i=a]=\E[Y_i(a)]$ and thus enables unbiased ATE estimates in RCT scenarios. 

\vspace{0.35\baselineskip}
\noindent \textbf{Remark}. The key assumption here, $\hat{Y}_i \perp A_i \;|\; Y_i$, indicates that, conditional on the true value of $Y_i$, there is no additional signal relating $A_i$ and $\hat{Y}_i$. For example, if $A_i$ is gender and $Y_i$ is income, the assumption states that, among truly high-income individuals, the model error does not predict gender. This unbiasedness result does therefore not mean all forms of bias are eliminated, simply that the bias added from attenuation is eliminated. See  Appendix I for a discussion of plug-in consistency, which is, in practice, complicated by how overfitting in the training set can lead to distribution shifts affecting performance. 

Viewed through an optimization lens, the Tweedie pseudo-outcome \(\widetilde Y_i = \hat Y_i - \sigma^2 \frac{d}{d \hat Y} \log p_{\hat Y}(\hat Y_i)\) is exactly a single gradient-descent step on the log-density \(\log p_{\hat Y}\), with learning rate \(\eta=\sigma^2\), that nudges predictions away from the over-shrunk high-density region toward the tails (see Appendix I). This perspective naturally points to a broader class of Tweedie-style updates obtained by varying the step size, number of steps, and implicit error model. In the Appendix, we discuss how one might organize the resulting \emph{Tweedie corrections} as a family indexed by the implicit error model that links the latent outcome \(Y_i\) and the ML prediction \(\hat Y_i\). The variants differ by the structure one is willing to assume and by how one wishes to correct global scaling (via a single calibration slope \(k\) and number of update steps) in addition to local shrinkage. In our experiments below, we focus on the Berkson variant we have been analyzing---leaving to future work elucidation of the full range of bias-variance trade-offs in the Tweedie family.

%% file: tweedie_proofs.tex

\begin{proposition}[Shift invariance]
\label{prop:shift_invariance}
Let $Z_i:=\hat Y_i+m$ for any constant $m$. Then the Berkson pseudo‑outcome computed from $Z_i$ equals
\(
\widetilde Z_i \,=\, m + \widetilde Y_i.
\)
Hence, any contrast (e.g., ATE) formed from $\widetilde Y_i$ is invariant to intercept shifts.
\emph{Proof.}
$s_Z(z)=\frac{d}{d z}\log p_Z(z)=\frac{d}{d z}\log p_{\hat Y}(z-m)=s_{\hat Y}(z-m)$.
Evaluate at $z=Z_i$ and expand $\widetilde Z_i$.
\end{proposition}

\begin{proposition}[Plug‑in consistency for Tweedie ATE]
\label{prop:plugin_consistency}
Suppose the score function, $\hat s_n$ and $\hat\sigma_n^2$ are estimated on an independent upstream sample partition and satisfy
$\E[(\hat s_n(\hat Y)-s_{\hat Y}(\hat Y))^2]\to 0$ and $\hat\sigma_n^2\to\sigma^2$ in probability.
Define $\widetilde Y_{i,n}:=\hat Y_i-\hat\sigma_n^2\,\hat s_n(\hat Y_i)$ on the downstream trial.
Then
\[
\hat\tau_n
\;:=\;
\hat{\E}[\widetilde Y_{i,n}\mid A_i=1]
-\hat{\E}[\widetilde Y_{i,n}\mid A_i=0]
\;\xrightarrow{p}\;
\tau,
\]
under random assignment and standard moment conditions.
\end{proposition}
\emph{Proof sketch.}
Use Proposition~\ref{prop:berkson_ate} for the oracle $(s_{\hat Y},\sigma^2)$,
then apply Slutsky’s theorem and a Law of Large Numbers with sample‑splitting to control plug‑in error.

\noindent \textbf{Remark.} In practice, plug-in consistency results may be compromised by the mechanics of ML model training, where training and inference set prediction distributions often, in finite datasets, show markedly different error dynamics (with training error collapsing to 0 while inference error possibly increases). This fact can induce forms of distribution shift in $\hat{Y}_i$ even if data at training and inference time are from the same distribution. That said, as $n\to\infty$, the concern of overfitting would be reduced with many model classes \citep{devroye2013probabilistic}.

\paragraph{Optimization view: Tweedie as a one–step gradient method.} 
Under the Berkson model, the pseudo-outcome in~\eqref{eq:po-berkson} is exactly a single explicit gradient step. Let $\phi(y):=\log p_{\hat Y}(y)$ denote the log-density, so that $\nabla\phi(y)=-
\frac{d}{d \hat{y}} \log p_{\hat Y_i}
$. Then
\[
\widetilde Y_i
= \hat Y_i - \underbrace{\sigma^2}_{\text{learning rate }\eta}\nabla\phi(\hat Y_i)
= \hat Y_i - \sigma^2 \frac{d}{d \hat{y}} \log p_{\hat Y_i},
\]
i.e., one step of \emph{gradient descent} on $\log p_{\hat Y}$ starting at $\hat Y_i$ with step size $\eta=\sigma^2$ and number of steps $t=1$. This makes sense for debiasing because $\hat{Y}_i$ values are shrunk toward the mode of $p_{\hat{Y}}$ (high-density region); descent moves them toward lower-density regions. Iterating this step would follow the gradient flow $y'(t)=\sigma^2\nabla\phi(y(t))$, but our method sets $t=1$ to avoid overshoot and contain variance, hinting at how there may be, indeed, a family of Tweedie corrections, each implying different performance dynamics.

\subsection*{Error Frameworks: A  Family of Tweedie Corrections?}

We can consider the classical conceptual model for measurement error,
\[
\hat{Y}_i = k Y_i + \epsilon_i \;\; \textrm{ with } \epsilon_i 
\overset{\text{iid}}{\sim}
 N(0,\;\sigma^2).
\]
or the Berkson-style model as a debiasing perspective \citep{carroll2006measurement,heid2004two}, used in the main body: 
\[
Y_i = k \hat{Y}_i + \epsilon_i. 
\]

In the former (the classical model), predictions \(\hat{Y}_i\) are attenuated (via \(k < 1\)) and noisy versions of the true outcomes \(Y_i\), pulling the extremes toward the mean due to regularization or intrinsic noise. In the latter (the Berkson-error variant), the relationship is inverted, implying truths are scaled predictions plus noise. What are the implications of these alternative models?

Recall the canonical formulation of Tweedie's identity: 
\begin{itemize}
\item[] ``For any random variable $Z$, 
\\ where $Z=\omega+\varepsilon$ with $\varepsilon\sim\mathcal N(0,\sigma^{2})$, 
\item[] \;\; $\E[\omega\mid Z=z]=z+\sigma^{2}\,\frac{d}{dz}\log p_Z(z)$.'' 
\end{itemize}
Under the classical model, we can set \(W_i = k Y_i\), so \(\hat{Y}_i = W_i + \epsilon_i\); then Tweedie gives \(\mathbb{E}[W_i \mid \hat{Y}_i = \hat{y}] = \hat{y} + \sigma^2 \frac{d}{d\hat{y}} \log p_{\hat{Y}}(\hat{y})\), and divide by \(k\) post-adjustment to recover \(\mathbb{E}[Y_i \mid \hat{Y}_i]\). This approach conditions on the observed \(\hat{Y}_i\) (available downstream) to estimate latent \(Y_i\), directly yielding debiased causal estimates. 

When \(k\neq 1\) (in either the classical or Berkson view), \(k\) and \(\sigma\) must be jointly calibrated: changing \(k\) alters the implied error scale (through the residuals used to estimate \(\sigma^2\)). To sidestep these joint model complexities for now, one could focus on $k=1$ variants. Because the classical $k=1$ variant cannot model shrinkage (as the additive noise means that the $\hat{Y}_i$ has greater variance than $Y_i$, which is undesirable for the model approach here), we therefore in this work focus on the Berkson $k=1$ setting.

The Berkson approach requires deploying a more complicated score-swapping schema than under the classical model. A direct application of Tweedie's formula yields (letting $k=1$): 
\begin{align*} 
\mathbb{E}[\hat{Y}_i \mid Y_i = y] = y + \sigma^2 \frac{d}{d  y} \log p_{Y}(y).
\end{align*} 
However, we want $\mathbb{E}[Y_i \mid \hat{Y}_i = y]$. So, we have some more work to do. First, we note that rearrangement yields: 
\begin{align*} 
y = \mathbb{E}[\hat{Y}_i \mid Y_i = y] - \sigma^2 \frac{d}{d  y} \log p_{Y}(y)
\end{align*} 
Under further assumptions (see below discussion of score swapping), we find 
\begin{equation}
    \frac{d}{dy} \log p_Y(y) = \mathbb{E} \left[ \frac{d}{d\hat{Y}} \log p_{\hat{Y}}(\hat{Y}) \mid Y_i = y \right],\label{eq:score_y_is_e_score_hat}
\end{equation}
yielding  
\begin{align*} 
y = \mathbb{E}[\hat{Y}_i \mid Y_i = y] - \sigma^2 \mathbb{E}\left[ \frac{d}{d\hat{Y}} \log p_{\hat{Y}}(\hat{Y}) \mid Y_i = y \right].
\end{align*} 
Hence, by linearity of conditional expectation, we get
\begin{align*} 
y = \mathbb{E}\left[\hat{Y}_i - \sigma^2  \frac{d}{d\hat{y}} \log p_{\hat{Y}}(\hat{Y}_i) \mid Y_i = y \right],
\end{align*} 
yielding the plug-in pseudo-outcome motivated by the conditional-expectation identity:
$$
\tilde{Y}_i = \hat{Y}_i - \sigma^2  \frac{d}{d\hat{y}} \log p_{\hat{Y}}(\hat{Y}_i)
$$
Under attenuation, $y<\mathbb{E}[\hat{Y}_i\mid y]$ for values of $y$ below the central mass, and  $y>\mathbb{E}[\hat{Y}_i\mid y]$ for values of $y$ above.
As $\mathbb{E}[\tilde{Y}_i | y] = y$ for all $y$, we have now eliminated this attenuation effect, though note that other biases may remain.

\subsubsection{Score Swapping}
\label{sec:score_swapping}

Consider the Berkson-variant error model:
$$
Y_i = \hat{Y}_i + \epsilon_i, \quad \epsilon_i \perp \hat{Y}_i, \quad \epsilon_i \sim \mathcal{N}(0, \sigma^2)
$$
From Tweedie's formula,
$$
\mathbb{E}[\hat{Y} \mid Y = y] = y + \sigma^2 \frac{d}{dy} \log p_Y(y)
$$
Let $p_Y$ denote the marginal density of $Y$ and $p_{\hat{Y}}$ the marginal density of $\hat{Y}$. We wish to show that
$$
\frac{d}{dy} \log p_Y(y) = \mathbb{E} \left[ \frac{d}{d\hat{y}} \log p_{\hat{Y}}(\hat{Y}) \mid Y = y \right]
$$

\begin{figure}[htb]
    \centering
    \includegraphics[width=1\linewidth]{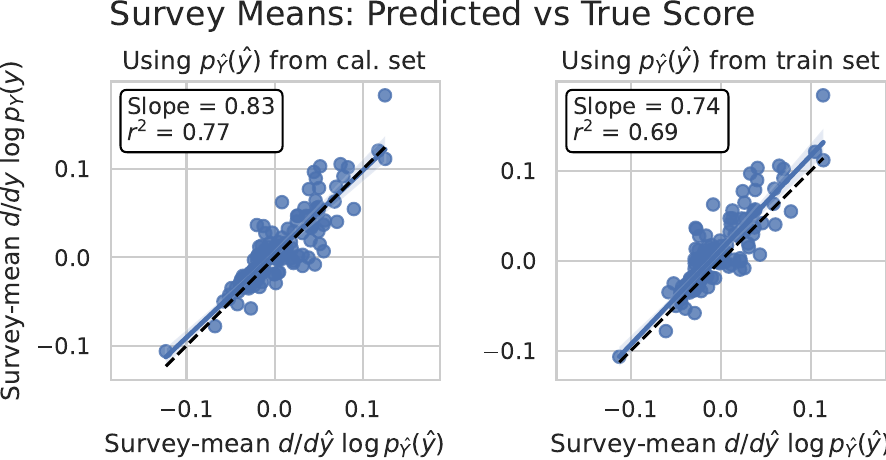}
    \caption{
Correlation between the score functions \(\frac{d}{dy} \log p_{Y}(y)\) and \(\frac{d}{d\hat{y}} \log p_{\hat{Y}}(\hat{y})\), showing the mean true score (from \(p_Y(y)\)) vs. the mean predicted score (from \(p_{\hat{Y}}(\hat{y})\)) for each survey. Score functions estimated from both the held-out calibration set and the training set demonstrate strong correlation with the true score.}
\label{fig:score_swapping_res}
\end{figure}

Since $Y = \hat{Y} + \epsilon$ with $\hat{Y} \perp \epsilon$, the marginal density of $Y$ can be expressed as the convolution
\begin{equation}
    p_Y(y) = \int p_{\hat{Y}}(\hat{y})p_{\epsilon}(y - \hat{y}) \, d\hat{y}
\end{equation}
To compute $\frac{d}{dy} \log p_Y (y)$, first differentiate $p_Y(y)$ itself. Under mild regularity conditions allowing differentiation under the integral sign (satisfied for the Gaussian noise and typical priors),
\begin{align*}
    p'_Y(y) &= \frac{d}{dy} \int p_{\hat{Y}}(\hat{y}) p_{\epsilon}(y - \hat{y}) \, d\hat{y}\\
    &= \int p_{\hat{Y}}(\hat{y}) \frac{d}{dy} p_{\epsilon}(y - \hat{y}) \, d\hat{y}
\end{align*}
Note that for the argument $y - \hat{y}$,
$$
\frac{d}{dy} p_{\epsilon}(y - \hat{y}) = - \frac{d}{d\hat{y}} p_{\epsilon}(y - \hat{y})
$$
Therefore,
$$
p'_{Y}(y) = - \int p_{\hat{Y}}(\hat{y}) \frac{d}{d\hat{y}} p_{\epsilon}(y - \hat{y}) \, d\hat{y}
$$
Applying integration by parts in $\hat{y}$,
$$
p'_Y(y) = \left[ - p_{\hat{Y}}(\hat{y})p_{\epsilon}(y - \hat{y}) \right]_{-\infty}^{\infty} + \int p_{\epsilon}(y - \hat{y}) \frac{d}{d\hat{y}} p_{\hat{Y}}(\hat{y}) \, d\hat{y}
$$
Under mild tail conditions (satisfied for the Gaussian noise), the boundary term vanishes. Using the identity $p'_{\hat{Y}}(\hat{y}) = p_{\hat{Y}}(\hat{y}) \frac{d}{d\hat{y}} \log p_{\hat{Y}}(\hat{y})$,
$$
p'_{Y}(y) = \int p_{\epsilon}(y - \hat{y}) p_{\hat{Y}}(\hat{y}) \frac{d}{d\hat{y}} \log p_{\hat{Y}}(\hat{y}) \, d\hat{y}
$$
Using the identity
$$
\frac{d}{dy} \log p_Y (y) = \frac{p'_Y(y)}{p_Y(y)}
$$
divide by $p_Y(y)$
\begin{align}
    \frac{d}{dy} \log p_Y (y) &= \frac{\int p_{\epsilon}(y - \hat{y}) p_{\hat{Y}}(\hat{y}) \frac{d}{d\hat{y}} \log p_{\hat{Y}}(\hat{y}) \, d\hat{y}}{p_Y(y)} \\
    & = \int \frac{p_{\hat{Y}}(\hat{y}) p_{\epsilon}(y - \hat{y})}{p_Y(y)} \frac{d}{d\hat{y}} \log p_{\hat{Y}}(\hat{y}) \, d\hat{y}
    \label{eq:d_dy_log_py}
\end{align}
From Bayes' rule, we have that
$$
p_{\hat{Y} \mid Y}(\hat{y} \mid y) = \frac{p_{\hat{Y}}(\hat{y}) p_{\epsilon}(y - \hat{y})}{p_Y(y)}
$$
Substituting this into \ref{eq:d_dy_log_py}, the ratio becomes
\begin{align*}
    \frac{d}{dy} \log p_Y (y) &= \int \frac{p_{\hat{Y}}(\hat{y}) p_{\epsilon}(y - \hat{y})}{p_Y(y)} \frac{d}{d\hat{y}} \log p_{\hat{Y}}(\hat{y}) \, d\hat{y} \\
    &= \int p_{\hat{Y} \mid Y}(\hat{y} \mid y) \frac{d}{d\hat{y}} \log p_{\hat{Y}}(\hat{y}) \, d\hat{y} \\
\end{align*}
This integral is precisely the conditional expectation of the score of $p_{\hat{Y}}$ given $Y=y$. Thus,
$$
\boxed{
\frac{d}{dy} \log p_Y(y) = \mathbb{E} \left[ \frac{d}{d\hat{Y}} \log p_{\hat{Y}}(\hat{Y}) \mid Y = y \right].
}
$$
An empirical evaluation of this relationship on our poverty prediction dataset is presented in Figure \ref{fig:score_swapping_res}.

%% file: main.bbl
\begin{thebibliography}{33}
\providecommand{\natexlab}[1]{#1}

\bibitem[{Angelopoulos et~al.(2023)Angelopoulos, Bates, Fannjiang, Jordan, and
  Zrnic}]{angelopoulos2023prediction}
Angelopoulos, A.~N.; Bates, S.; Fannjiang, C.; Jordan, M.~I.; and Zrnic, T.
  2023.
\newblock Prediction-powered Inference.
\newblock \emph{Science}, 382(6671): 669--674.

\bibitem[{Berglund(2012)}]{berglund2012regression}
Berglund, L. 2012.
\newblock Regression dilution bias: tools for correction methods and sample
  size calculation.
\newblock \emph{Upsala journal of medical sciences}, 117(3): 279--283.

\bibitem[{Burke et~al.(2021)Burke, Driscoll, Lobell, and
  Ermon}]{burkeUsingSatelliteImagery2021}
Burke, M.; Driscoll, A.; Lobell, D.~B.; and Ermon, S. 2021.
\newblock Using Satellite Imagery to Understand and Promote Sustainable
  Development.
\newblock 371(6535): eabe8628.

\bibitem[{Carroll et~al.(2006)Carroll, Ruppert, Stefanski, and
  Crainiceanu}]{carroll2006measurement}
Carroll, R.~J.; Ruppert, D.; Stefanski, L.~A.; and Crainiceanu, C.~M. 2006.
\newblock \emph{{Measurement Error in Nonlinear Models: A Modern Perspective}}.
\newblock Chapman and Hall/CRC.

\bibitem[{Chi et~al.(2022)Chi, Fang, Chatterjee, and
  Blumenstock}]{chi2022microestimates}
Chi, G.; Fang, H.; Chatterjee, S.; and Blumenstock, J.~E. 2022.
\newblock Microestimates of wealth for all low-and middle-income countries.
\newblock \emph{Proceedings of the National Academy of Sciences}, 119(3):
  e2113658119.

\bibitem[{Conlin(2024)}]{Conlin2024}
Conlin, C. 2024.
\newblock \emph{Using Machine Learning and Daytime Satellite Imagery to
  Estimate Aid's Effect on Wealth: Comparing China and World Bank Programs in
  Africa}.
\newblock Master's thesis, Linkoping University, Department of Management and
  Engineering, The Institute for Analytical Sociology (IAS), Linkoping, Sweden.
\newblock URN: urn:nbn:se:liu:diva-205256, ISRN: LIU-IEI-FIL-A--24/04656--SE.

\bibitem[{Daoud and Dubhashi(2023)}]{daoudStatisticalModelingThree2023}
Daoud, A.; and Dubhashi, D. 2023.
\newblock Statistical {{Modeling}}: {{The Three Cultures}}.
\newblock 5(1).

\bibitem[{Daoud and Johansson(2024)}]{daoudImpactAusterityChildren2024}
Daoud, A.; and Johansson, F.~D. 2024.
\newblock The Impact of Austerity on Children: {{Uncovering}} Effect
  Heterogeneity by Political, Economic, and Family Factors in Low- and
  Middle-Income Countries.
\newblock 118: 102973.

\bibitem[{Daoud et~al.(2023)Daoud, Jord{\'a}n, Sharma, Johansson, Dubhashi,
  Paul, and Banerjee}]{daoud2023using}
Daoud, A.; Jord{\'a}n, F.; Sharma, M.; Johansson, F.; Dubhashi, D.; Paul, S.;
  and Banerjee, S. 2023.
\newblock Using satellite images and deep learning to measure health and living
  standards in India.
\newblock \emph{Social Indicators Research}, 167(1): 475--505.

\bibitem[{de~Pieuchon et~al.(2025)de~Pieuchon, Daoud, Jerzak, Johansson, and
  Johansson}]{depieuchon2025benchmarkingdebiasingmethodsllmbased}
de~Pieuchon, N.~A.; Daoud, A.; Jerzak, C.~T.; Johansson, M.; and Johansson, R.
  2025.
\newblock Benchmarking Debiasing Methods for LLM-based Parameter Estimates.
\newblock arXiv:2506.09627.

\bibitem[{Devroye, Gy{\"o}rfi, and Lugosi(2013)}]{devroye2013probabilistic}
Devroye, L.; Gy{\"o}rfi, L.; and Lugosi, G. 2013.
\newblock \emph{{A Probabilistic Theory of Pattern Recognition}}, volume~31.
\newblock Springer Science \& Business Media.

\bibitem[{Efron(2011)}]{efron2011tweedie}
Efron, B. 2011.
\newblock Tweedie’s Formula and Selection Bias.
\newblock \emph{Journal of the American Statistical Association}, 106(496):
  1602--1614.

\bibitem[{Egami et~al.(2023)Egami, Hinck, Stewart, and Wei}]{dsl_original}
Egami, N.; Hinck, M.; Stewart, B.; and Wei, H. 2023.
\newblock Using Imperfect Surrogates for Downstream Inference: Design-based
  Supervised Learning for Social Science Applications of Large Language Models.
\newblock In Oh, A.; Naumann, T.; Globerson, A.; Saenko, K.; Hardt, M.; and
  Levine, S., eds., \emph{Advances in Neural Information Processing Systems},
  volume~36, 68589--68601. Curran Associates, Inc.

\bibitem[{Grimmer, Roberts, and Stewart(2022)}]{grimmer2022text}
Grimmer, J.; Roberts, M.~E.; and Stewart, B.~M. 2022.
\newblock \emph{Text as data: A new framework for machine learning and the
  social sciences}.
\newblock Princeton University Press.

\bibitem[{Heid et~al.(2004)Heid, K{\"u}chenhoff, Miles, Kreienbrock, and
  Wichmann}]{heid2004two}
Heid, I.; K{\"u}chenhoff, H.; Miles, J.; Kreienbrock, L.; and Wichmann, H.
  2004.
\newblock {Two Dimensions of Measurement Error: Classical and Berkson Error in
  Residential Radon Exposure Assessment}.
\newblock \emph{Journal of Exposure Science \& Environmental Epidemiology},
  14(5): 365--377.

\bibitem[{Jean et~al.(2016)Jean, Burke, Xie, Davis, Lobell, and
  Ermon}]{jean2016combining}
Jean, N.; Burke, M.; Xie, M.; Davis, W.~M.; Lobell, D.~B.; and Ermon, S. 2016.
\newblock {Combining Satellite Imagery and Machine Learning to Predict
  Poverty}.
\newblock \emph{Science}, 353(6301): 790--794.

\bibitem[{Jerzak, Johansson, and
  Daoud(2023{\natexlab{a}})}]{jerzak2023integrating}
Jerzak, C.~T.; Johansson, F.; and Daoud, A. 2023{\natexlab{a}}.
\newblock Integrating earth observation data into causal inference: challenges
  and opportunities.
\newblock \emph{arXiv preprint arXiv:2301.12985}.

\bibitem[{Jerzak, Johansson, and Daoud(2023{\natexlab{b}})}]{jerzak2023image}
Jerzak, C.~T.; Johansson, F.~D.; and Daoud, A. 2023{\natexlab{b}}.
\newblock Image-based Treatment Effect Heterogeneity.
\newblock In \emph{Conference on Causal Learning and Reasoning}, 531--552.
  PMLR.

\bibitem[{Kakooei et~al.(2024)Kakooei, Bailie, S{\"o}derberg, Becevic, and
  Daoud}]{kakooei2024mapping}
Kakooei, M.; Bailie, J.; S{\"o}derberg, A.; Becevic, A.; and Daoud, A. 2024.
\newblock Mapping Africa Settlements: High Resolution Urban and Rural Map by
  Deep Learning and Satellite Imagery.
\newblock \emph{arXiv preprint arXiv:2411.02935}.

\bibitem[{Kino et~al.(2021)Kino, Hsu, Shiba, Chien, Mita, Kawachi, and
  Daoud}]{kino2021scoping}
Kino, S.; Hsu, Y.-T.; Shiba, K.; Chien, Y.-S.; Mita, C.; Kawachi, I.; and
  Daoud, A. 2021.
\newblock A scoping review on the use of machine learning in research on social
  determinants of health: Trends and research prospects.
\newblock \emph{SSM-population Health}, 15: 100836.

\bibitem[{Lu, Bates, and Wang(2024)}]{lu2024quantifying}
Lu, K.; Bates, S.; and Wang, S. 2024.
\newblock {Quantifying Uncertainty in Area and Regression Coefficient
  Estimation from Remote Sensing Maps}.
\newblock \emph{arXiv preprint arXiv:2407.13659}.

\bibitem[{Malik et~al.(2021)Malik, Parks, Russell, Lin, Walsh, Solomon, Zhang,
  Elston, and Goodman}]{malik2021banking}
Malik, A.; Parks, B.; Russell, B.; Lin, J.; Walsh, K.; Solomon, K.; Zhang, S.;
  Elston, T.; and Goodman, S. 2021.
\newblock Banking on the Belt and Road: Insights from a new global dataset of
  13,427 Chinese development projects.
\newblock \emph{Williamsburg, VA: AidData at William \& Mary}, 23--36.

\bibitem[{Olofsson et~al.(2013)Olofsson, Foody, Stehman, and
  Woodcock}]{olofsson2013making}
Olofsson, P.; Foody, G.~M.; Stehman, S.~V.; and Woodcock, C.~E. 2013.
\newblock {Making Better Use of Accuracy Data in Land Change Studies:
  Estimating Accuracy and Area and Quantifying Uncertainty Using Stratified
  Estimation}.
\newblock \emph{Remote sensing of environment}, 129: 122--131.

\bibitem[{Pettersson et~al.(2023)Pettersson, Kakooei, Ortheden, Johansson, and
  Daoud}]{pettersson2023time}
Pettersson, M.~B.; Kakooei, M.; Ortheden, J.; Johansson, F.~D.; and Daoud, A.
  2023.
\newblock Time Series of Satellite Imagery Improve Deep Learning Estimates of
  Neighborhood-Level Poverty in Africa.
\newblock In \emph{IJCAI}, 6165--6173.

\bibitem[{Ratledge et~al.(2022)Ratledge, Cadamuro, de~la Cuesta, Stigler, and
  Burke}]{ratledge_using_2022}
Ratledge, N.; Cadamuro, G.; de~la Cuesta, B.; Stigler, M.; and Burke, M. 2022.
\newblock {Using Machine Learning to Assess the Livelihood Impact of
  Electricity Access}.
\newblock \emph{Nature}, 611(7936): 491--495.
\newblock Number: 7936 Publisher: Nature Publishing Group.

\bibitem[{Robbins(1956)}]{robbins1956empirical}
Robbins, H. 1956.
\newblock An Empirical Bayes Approach to Statistics.
\newblock In \emph{Proceedings of the Third Berkeley Symposium on Mathematical
  Statistics and Probability, 1954--1955}, volume~I, 157--163. Berkeley and Los
  Angeles: University of California Press.

\bibitem[{Sakamoto, Jerzak, and Daoud(2025)}]{sakamoto2024scoping}
Sakamoto, K.; Jerzak, C.~T.; and Daoud, A. 2025.
\newblock A Scoping Review of Earth Observation and Machine Learning for Causal
  Inference: Implications for the Geography of Poverty.
\newblock In Hall, O.; and Wahab, I., eds., \emph{Geography of Poverty}.

\bibitem[{Shalev-Shwartz and Ben-David(2014)}]{shalev2014understanding}
Shalev-Shwartz, S.; and Ben-David, S. 2014.
\newblock \emph{{Understanding Machine Learning: From Theory to Algorithms}}.
\newblock Cambridge university press.

\bibitem[{Shu and Yi(2019)}]{shu2019causal}
Shu, D.; and Yi, G.~Y. 2019.
\newblock {Causal Inference with Measurement Error in Outcomes: Bias Analysis
  and Estimation Methods}.
\newblock \emph{Statistical methods in medical research}, 28(7): 2049--2068.

\bibitem[{Stewart et~al.(2023)Stewart, Lehmann, Corley, Wang, Chang, Braham,
  Sehgal, Robinson, and Banerjee}]{stewart2023ssl4eoldatasetsfoundationmodels}
Stewart, A.~J.; Lehmann, N.; Corley, I.~A.; Wang, Y.; Chang, Y.-C.; Braham, N.
  A.~A.; Sehgal, S.; Robinson, C.; and Banerjee, A. 2023.
\newblock SSL4EO-L: Datasets and Foundation Models for Landsat Imagery.
\newblock arXiv:2306.09424.

\bibitem[{Ting(2024{\natexlab{a}})}]{ting2024machine}
Ting, Y.-S. 2024{\natexlab{a}}.
\newblock Why Machine Learning Models Systematically Underestimate Extreme
  Values.
\newblock \emph{arXiv preprint arXiv:2412.05806}.

\bibitem[{Ting(2024{\natexlab{b}})}]{ting2024machinelearningmodelssystematically}
Ting, Y.-S. 2024{\natexlab{b}}.
\newblock Why Machine Learning Models Systematically Underestimate Extreme
  Values.
\newblock arXiv:2412.05806.

\bibitem[{Zhu, Jerzak, and Daoud(2025)}]{pmlr-v275-zhu25a}
Zhu, F.~W.; Jerzak, C.~T.; and Daoud, A. 2025.
\newblock Optimizing Multi-Scale Representations to Detect Effect Heterogeneity
  Using Earth Observation and Computer Vision: Applications to Two Anti-Poverty
  RCTs.
\newblock In Huang, B.; and Drton, M., eds., \emph{Proceedings of the Fourth
  Conference on Causal Learning and Reasoning}, volume 275 of \emph{Proceedings
  of Machine Learning Research}, 894--919. PMLR.

\end{thebibliography}
